\documentclass[sigconf]{acmart}
\usepackage{ulem}
\usepackage{float}
\usepackage{algorithm}
\usepackage{algpseudocode}
\usepackage[marginal]{footmisc}

\usepackage{graphics} 
\usepackage{epsfig} 
\usepackage{indentfirst}
\usepackage{amsmath}
\usepackage{bbding}
\usepackage{booktabs}
\usepackage{url}
\usepackage{multirow}
\usepackage{bbding}
\usepackage{hyperref}
\usepackage{blindtext}
\renewcommand\footnotetextcopyrightpermission[1]{} 
\AtBeginDocument{%
  \providecommand\BibTeX{{%
    \normalfont B\kern-0.5em{\scshape i\kern-0.25em b}\kern-0.8em\TeX}}}

\begin{document}

\title{WaterVG: Waterway Visual Grounding based on Text-Guided Vision and mmWave Radar}

\author{Runwei Guan$^{1,3,2}$,\hspace{0.3em} Liye Jia$^{1,3,2}$,\hspace{0.3em} Fengyufan Yang$^{2,1,3}$,\hspace{0.3em} Shanliang Yao$^{2,1,3}$, \hspace{0.3em} Erick Purwanto$^{2}$, \hspace{0.3em} Xiaohui Zhu$^{2}$, \hspace{0.3em} Eng Gee Lim$^{2}$, \hspace{0.3em} Jeremy Smith$^{3}$,\hspace{0.3em} Ka Lok Man$^{2}$, \hspace{0.3em} Xuming Hu$^{4}$, \hspace{0.3em} Yutao Yue$^{1,4*}$  }
\affiliation{%
\institution{1. Institute of Deep Perception Tech, JITRI;\hspace{0.3em} 2. School of Advanced Technology, Xi'an Jiaotong-Liverpool University \\
\{erick.purwanto,\hspace{0.2em}xiaohui.zhu,\hspace{0.2em}ka.man,\hspace{0.2em}enggee.lim\}@xjtlu.edu.cn
} 
\country{China}
}

%
\affiliation{%
\institution{3. Faculty of Science and Engineering, University of Liverpool \hspace{0.3em} \\
\{runwei.guan,\hspace{0.2em}liye.jia,\hspace{0.2em}fengyufan.yang,\hspace{0.2em}shanliang.yao,\hspace{0.2em}J.S.Smith\}@liverpool.ac.uk
} 
{\country{United Kingdom}}
}

\affiliation{%
\institution{4. Thrust of Artificial Intelligence and Thrust of Intelligent Transportation, HKUST (Guangzhou) \\
\{xuminghu, yutaoyue\}@hkust-gz.edu.cn
} 
\country{China}
}


\renewcommand{\shortauthors}{Guan et~al.}

\begin{abstract}
  The perception of waterways based on human intent is significant for autonomous navigation and operations of Unmanned Surface Vehicles (USVs) in water environments. Inspired by visual grounding, we introduce WaterVG, the first visual grounding dataset designed for USV-based waterway perception based on human prompts. WaterVG encompasses prompts describing multiple targets, with annotations at the instance level including bounding boxes and masks. Notably, WaterVG includes 11,568 samples with 34,987 referred targets, whose prompts integrates both visual and radar characteristics. The pattern of text-guided two sensors equips a finer granularity of text prompts with visual and radar features of referred targets. Moreover, we propose a low-power visual grounding model, Potamoi, which is a multi-task model with a well-designed Phased Heterogeneous Modality Fusion (PHMF) mode, including Adaptive Radar Weighting (ARW) and Multi-Head Slim Cross Attention (MHSCA). Exactly, ARW extracts required radar features to fuse with vision for prompt alignment. MHSCA is an efficient fusion module with a remarkably small parameter count and FLOPs, elegantly fusing scenario context captured by two sensors with linguistic features, which performs expressively on visual grounding tasks. Comprehensive experiments and evaluations have been conducted on WaterVG, where our Potamoi archives state-of-the-art performances compared with counterparts. 
\end{abstract}







\maketitle


\begin{figure}
    \includegraphics[width=1.00\linewidth]{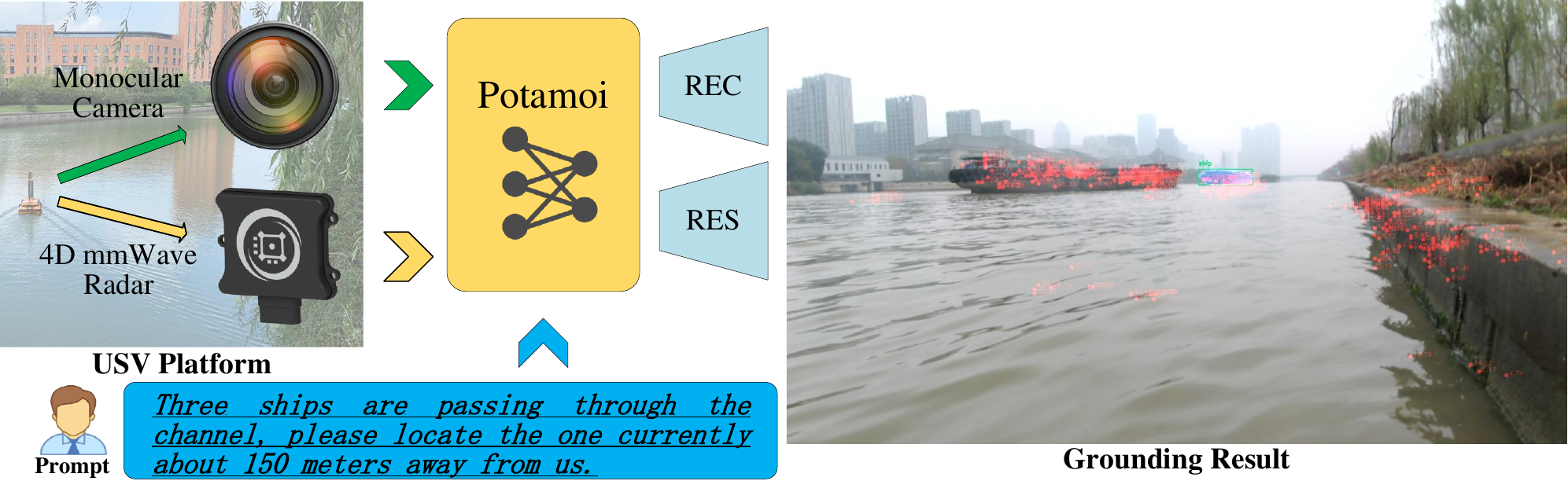}
    \vspace{-5mm}
    \caption{\textmd{Overview of our pipeline for WaterVG. The camera and radar align prompt on appearance, motion and distance features.}}
    \label{fig:overview}
\end{figure}

\section{Introduction}

With the advancement of deep learning, water-surface intelligent perception has witnessed substantial progress in various domains \cite{yaoreview2023,yao2023radar,yao2023waterscenes,guo2023asynchronous}. Specifically, current advancements primarily encompass object detection \cite{cheng2021flow,guan2023achelous++}, segmentation\cite{guan2023efficient,guan2023achelous,guan2024mask}, tracking \cite{zhang2021robust} and Simultaneous Localization And Mapping (SLAM) \cite{cheng2021we}. Based on the aforementioned, waterway perception faces the following challenges, \textbf{firstly}, waterways are often characterized by complexity, irregularity, and variability in obstacles and lighting conditions. \textbf{Secondly}, due to fluctuations in water levels and changes in flow velocity, many obstacles are irregular and shifting. \textbf{Thirdly}, vessel navigation in waterways usually does not follow prescribed routes, leading to a higher frequency of unforeseen events and increased challenges in waterway monitoring. Therefore, it is challenging for regulators or specialized USVs to locate desired waterway targets. 

However, current sensor-based full-target perception may not adequately address this issue. Hence, we pioneeringly introduce visual grounding to waterway perception, utilizing natural language as prompt to locate specific targets. Here, humans can convey target exact features through descriptions, facilitating more intuitive and efficient guidance for USVs. This leads to enhanced precision in navigation and localization, thereby holding significant implications for transportation, rescue operations, and environment conservation.

\begin{figure*}
    \includegraphics[width=0.97\linewidth]{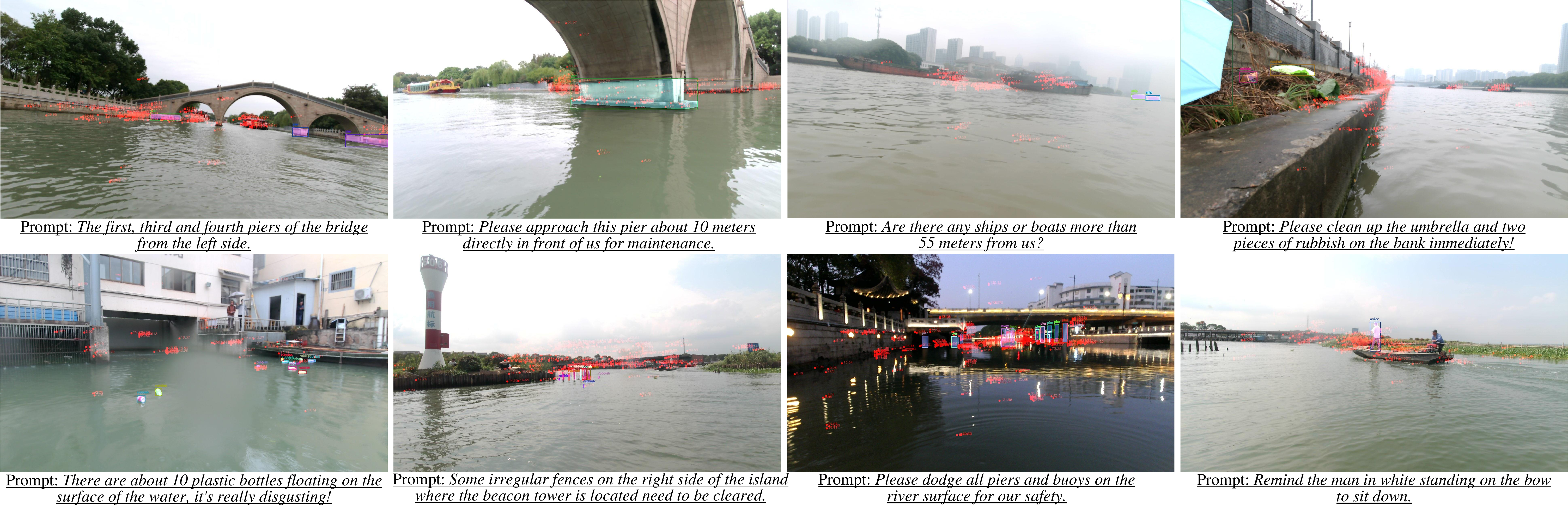}
    \vspace{-3mm}
    \caption{\textmd{Samples in WaterVG. The annotations include bounding boxes and masks, supplemented by radar point clouds (red).}}
    \label{fig:WaterVG_examples}
\end{figure*}

Building upon the aforementioned considerations, we contemplate the establishment of an optimal visual grounding dataset tailored for robust perception and monitoring in waterway environments. Here are our considerations, \textbf{firstly}, pure visual sensors are easily disturbed by variable visibility and lighting conditions \cite{cheng2021flow,yaoreview2023,yao2023waterscenes}. \textbf{Secondly}, visual sensors can only capture qualitative features of targets in terms of appearance or position, but traffic supervisors often need quantitative distance and motion information about waterway targets. Therefore, when one desires to ascertain whether there is a vessel approaching behind the fog or whether obstacles exist within a specific distance, visual sensors are unable to precisely locate targets based on human intent. To enable target referring flexibly, we complement the visual sensor with 4D mmWave radar, as it is an all-weather sensor that can detect targets in the form of point clouds and offer higher resolution than conventional mmWave radar \cite{guan2023achelous,guan2023achelous++,cheng2021robust}. Concurrently, while visual sensors perceive appearance features, including color, size, texture and shape, radar can furnish target reflection characteristics \cite{yao2023radar}, including distance, velocity, reflected power, etc. 

Upon these insights, we build the first visual grounding dataset called WaterVG for waterway perception, based on the WaterScenes dataset \cite{yao2023waterscenes}, which is a USV-based waterway perception dataset integrating the monocular camera and 4D radar. Detailedly, diverging from other vision-based visual grounding datasets, our WaterVG dataset has the following distinguishing features: \textbf{i)} Manual fine-grained prompts to describe targets through both visual attributes like color, size, shape, texture and features detectable by radar, such as numeric distance and motion characteristics. \textbf{ii)} Flexible referred targets with 1 to 15 per prompt. \textbf{iii)} Diverse prompts, containing specific descriptions, coarse and fine queries on single or multiple characteristics. \textbf{iv)} Multi-task annotations of Referring Expression Comprehension \cite{qiao2020referring} and Segmentation \cite{chrupala2022visually} (REC and RES).

Correspondingly, three heterogeneous modalities, RGB image, 3D point clouds and natural language will face several challenges regarding fusion for USV-based visual grounding. \textbf{Firstly}, radar may be interfered with clutter, causing some non-target points (ghost echoes). Hence, it is crucial to adaptively extract radar features mentioned in prompts while ignoring clutter. \textbf{Secondly}, referred targets vary significantly in visual and point cloud scale, so the dynamic fusion of sparse sensor features with corresponding dense language features is vital. \textbf{Thirdly}, fragmented feature extraction and fusion branches of multi-modal networks tend to cause increased power consumption and reduction in USV endurance time.

Therefore, we propose a low-power multi-task model called Potamoi, achieving a trade-off between accuracy and power consumption. Within this, we design an effective fusion mode called Phased Heterogeneous Modality Fusion (PHMF). PHMF performs a phased alignment and fusion of visual, radar, and linguistic features. In PHMF, Adaptive Radar Weighting (ARW) is proposed for high-quality radar feature extraction, effectively suppressing interference from clutter in space and adaptively extracting channel features of target radar characteristics contained within text prompts to the maximum extent. More importantly, PHMF's core lies in our proposed Multi-Head Slim Cross Attention (MHSCA). MHSCA efficiently reduces redundancy in query context and fuses contextual features captured by two sensors with linguistic features, achieving exemplary performance on visual grounding with lower parameters and FLOPs than vanilla Multi-Head Cross Attention (MHCA) \cite{wu2023referring} and Multi-Head Linear Cross Attention (MHLCA) \cite{choromanski2020rethinking}.


In summary, the contributions of this paper are as follows:

\begin{enumerate}
    \item The first visual grounding dataset for USV-based waterway perception, WaterVG, comprises 11,568 triplet input data, including images, radar point clouds, and prompts. Moreover, the ground truth encompasses two types of instance-level annotations: bounding boxes and masks. In particular, our textual prompts contain extra corresponding descriptions of target distance and motion features.
    \item A low-power multi-task visual grounding model, Potamoi, where an effective fusion mode called PHMF, aligns and fuses three modalities. PHMF includes ARW for radar feature adaptive extraction and a low-cost fusion method called MHSCA, which can fuse heterogeneous modalities efficiently for visual grounding with few parameters, low FLOPs and power than vanilla MHCA and MHLCA. 
    \item Comprehensive experiments based on the WaterVG dataset encompass various aspects such as the performance of visual grounding models under fusion and vision patterns, and multi-modal fusion methods. These experiments deepen insights and understanding of the field.
\end{enumerate}


\begin{figure*}
    \includegraphics[width=0.99\linewidth]{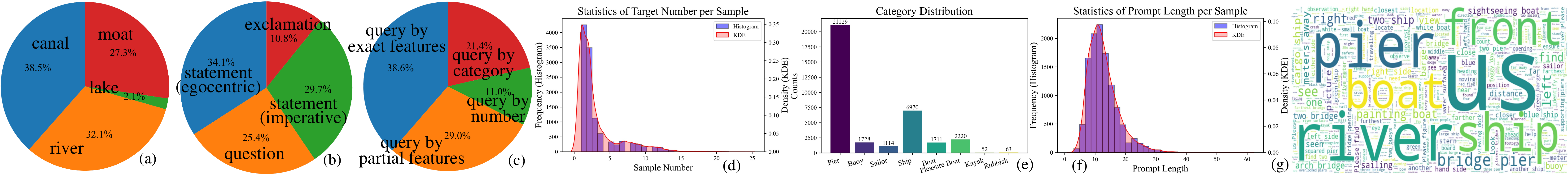}
    \vspace{-3mm}
    \caption{\textmd{The statistics of WaterVG, including proportion on the proportion of waterways (a), sentence patterns (b) and query types (c), and the statistics of WaterVG on referred target number (d), category distribution (e), prompt length (f) and word cloud (g).}}
    \label{fig:dataset_statistics_1}
\end{figure*}

\section{Related Works}
\label{sec:related}

\subsection{Waterway Perception Datasets}
Recently, waterway perception datasets have obtained significant attention in the field of waterway robotics and autonomous navigation. Most waterway perception datasets cover instance-level water-surface obstacle detection \cite{kristan2015fast,bovcon2018stereo,liu2021efficient,cheng2021flow}, segmentation \cite{bovcon2019mastr1325,taipalmaa2019high,zhou2021image,vzust2023lars} and tracking \cite{moosbauer2019benchmark,varga2022seadronessee} for traffic monitoring, autonomous driving and water environmental protection. Furthermore, datasets for the SLAM \cite{miller2018visual,cheng2021we} and segmentation \cite{yao2023waterscenes,cheng2021we} of drivable-area and shorelines are also proposed for the planning-oriented perception of USVs. Based on the above observations, it is evident that current waterway perception datasets primarily focus on sensor-based automatic perception, lacking flexibility in human-machine interaction. Text prompts can be regarded as a way to simulate human instructions. In many scenarios, leveraging text prompts, USVs (waterborne robots) can more flexibly locate and identify specific targets, thereby achieving more efficient execution of waterway tasks. Therefore, we design and construct the first waterway visual grounding dataset, WaterVG, to promote human-guided waterway monitoring and USV-based embodied AI in waterway perception.


\subsection{Visual Grounding Datasets and Models}
As a challenging branch within multi-modal learning, visual grounding aims to locate, based on text query, the targets in images that best match the descriptions provided by the natural language \cite{zhou2023vision}. This is further divided into REC and RES tasks. The REC task focuses on localizing targets with bounding boxes, whereas RES aims to delineate target masks. The majority of current 2D visual grounding datasets \cite{kazemzadeh2014referitgame,plummer2015flickr30k,yu2016modeling,krishna2017visual,deruyttere2019talk2car,wang2020give,chen2020cops,zhan2023rsvg,liu2023gres,wu2023referring,he2023grec} primarily cater to REC tasks, relying solely on pure visual data, lacking features from other synchronous modalities queried by prompts. Moreover, there is a current trend towards a prompt capable of querying multiple targets, which aligns more flexibly with real-world scenarios, deviating from the single-target reference. Moreover, visual grounding models are primarily categorized into three classes: two-stage \cite{liu2019learning,yang2019dynamic,yu2018mattnet,wang2019neighbourhood,hong2019learning}, one-stage \cite{luo2020multi,chen2018real,yang2019fast}, and transformer-based models \cite{zhu2022seqtr,du2022visual,deng2021transvg}. Among them, firstly, two-stage models split the generation of Regions Of Interest (ROI) and target classification into two parts, where the first stage is responsible for matching the regions of interest with text features while the second stage is to filter and refine ROI based on the fused features of text and image. Secondly, one-stage models usually combine bounding box regression and confidence prediction into one step. Besides, linguistic features usually fuse with image features at the end of the image backbone or within multi-scale fusion networks. Thirdly, transformer-based models are usually in the format of an image backbone and a text encoder followed by a transformer-based encoder and decoder, where the fusion operation usually happens before the transformer-based encoder.

\section{The WaterVG Dataset}
\label{sec:dataset}

Overall, the WaterVG dataset comprises 11,568 samples, including 9,254 in the training set, 578 in the validation set, and 1,736 in the test set. Each sample consists of an image, a frame of 4D radar point cloud, a prompt description, and ground truth presented in the forms of bounding boxes and masks. This section will sequentially elaborate on the construction process (Section \ref{subsec:construction}) and data statistics (Section \ref{subsec:statistics}) of the WaterVG dataset.

\begin{figure}
    \includegraphics[width=0.99\linewidth]{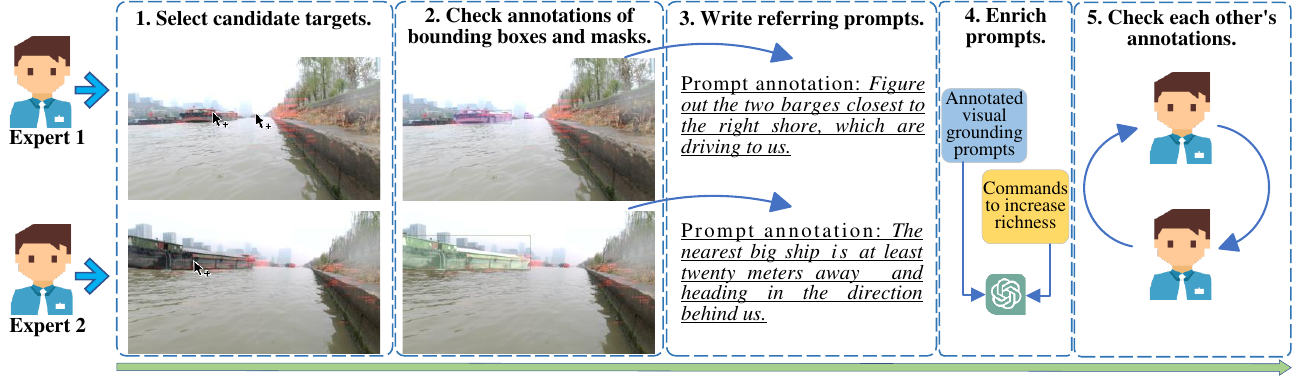}
    \vspace{-3mm}
    \caption{\textmd{Five-step prompt annotations of WaterVG.}}
    \label{fig:annotation_process}
\end{figure}

\begin{figure*}
    \includegraphics[width=0.99\linewidth]{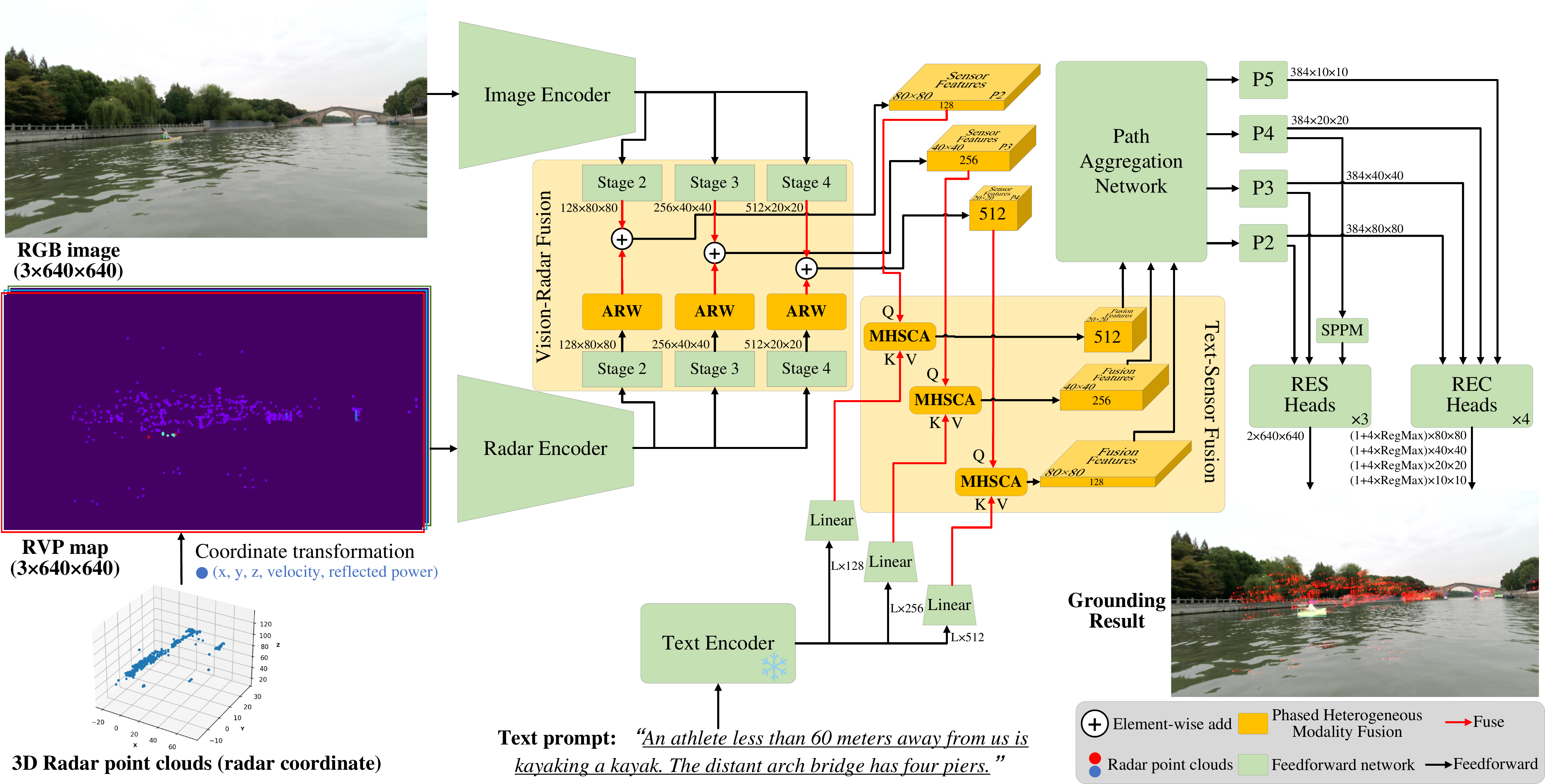}
    \vspace{-3mm}
    \caption{\textmd{The architecture of Potamoi. The text encoder (ALBERT) \cite{lan2020albert} for text prompts is frozen during training.}}
    \label{fig:main_model}
\end{figure*}

\subsection{Dataset Construction}
\label{subsec:construction}

Our WaterVG dataset is built upon the WaterScenes dataset \cite{yao2023waterscenes}, a high-quality, multi-sensor dataset with vision and radar, specifically designed for various water-surface environments. It features high-resolution images from a monocular camera with 1920 × 1080 pixels, complemented by 4D radar point clouds, which are calibrated and time-synchronized with the monocular camera. Additionally, WaterScenes \cite{yao2023waterscenes} provides high-quality annotations for object detection in the form of bounding boxes, as well as segmentation masks. Building upon these resources, we engage two experts with extensive experience in intelligent waterway transportation to conduct prompt annotation of our WaterVG dataset, selecting 11,568 representative samples from WaterScenes under different scenarios and lighting conditions. As illustrated in Fig. \ref{fig:annotation_process}, the annotation process is divided into five distinct stages.

\textbf{(1) Select candidate targets.} To facilitate efficient annotation by experts, we develop an annotation GUI client specifically for the WaterVG dataset. In this system, the radar point clouds are projected onto the camera plane using the projection matrix based on the extrinsic parameters of the radar and camera, as well as the intrinsic parameters of the camera, with the point clouds displayed in red. Each point cloud is accompanied by information displaying the distance of the target from the USV and its radial velocity. Correspondingly, both bounding box and mask annotations are mapped onto the image. Annotators simply need to click on the location of the target to display these two types of annotations.

\textbf{(2) Check the annotations.} Once the bounding box and mask of a clicked object are displayed on the image, two annotators proceed to check the accuracy of corresponding ground truth. If the ground truth is verified as correct, they continue the next step. If not, the target is re-annotated to ensure the precision of the dataset.

\textbf{(3) Write referring prompts.} Based on the images and radar point clouds, two annotators include as many physical characteristics of the targets as possible in the annotation prompt. These encompass the target's category, size, color, and shape features perceptible by visual means. Furthermore, if the target is detected by the radar precisely accompanied by dense projected point clouds, we use characteristics of point clouds to further describe the target's distance and motion trends from our USV. Additionally, if multiple targets are involved, annotators will also work on expressing their spatial location relationships. Based on these, the annotators aim to diversify the prompts by incorporating various sentence structures and methods of referencing the target.

\textbf{(4) Enrich prompts.} When facing identical scenes or targets in different samples, the limited vocabulary and subjective tone of annotators are unavoidable. To enhance the diversity of prompt content, we utilize the existing annotated prompts and feed them into ChatGPT-3.5-Turbo API \cite{ouyang2022training}, aiming to enrich the prompts by generating semantically equivalent yet more varied expressions.

\textbf{(5) Check each other's prompt annotations.} To avoid ambiguity in prompts, we exchange prompts written by two experts. Each expert reviews and examines prompts written by the other, respecting respective writing styles. In cases where there is ambiguity or misidentification in target references, the two experts engage in discussions and reach a consensus for necessary modifications.

\subsection{Dataset Statistics}
\label{subsec:statistics}

Figure \ref{fig:dataset_statistics_1} presents the statistics of WaterVG from perspectives of waterway, sentence pattern and query type. The waterway scenario proportion in WaterVG, includes canal, river, lake and moat. Besides, sentence patterns for prompts, including statement (egocentric), question, statement (imperative) and exclamation. Prompt type proportion contains 4 main types of query by exact features, partial features, number and category, respectively. Figure \ref{fig:dataset_statistics_1} also illustrates the distribution statistics of referred target number, prompt length per sample, referred category distribution and word cloud. In WaterVG, \textbf{firstly}, each text prompt refers to 1-15 targets, with the majority consisting of 1-2 targets. \textbf{Secondly}, The categories of referred target include piers, buoys, sailors, ships, boats, pleasure boats, kayaks and rubbishes, a total of 34,987 referred targets. \textbf{Thirdly}, the distribution of text prompt lengths, ranging from 2 to 40, encompass simple phrases and detailed sentences. \textbf{Fourthly}, the word cloud displays high-frequency words. The above demonstrates the diversity and comprehensiveness of our WaterVG dataset regarding waterway visual grounding, which is primarily reflected in the diversity of waterway scenes, richness of textual prompts in terms of content, patterns, and types, diversity of mentioned target categories, and flexibility in mentioned target number.

\section{Methods}
\label{sec:methods}

\subsection{Task Formulation}
Potamoi is a one-stage multi-task visual grounding model for two tasks, REC and RES. Given triple inputs of an image $I \in R^{3 \times H_I \times H_W}$, a frame of radar point clouds $RPC \in R^{5 \times N_R}$, containing the features $\{x, y, z, v, p\}$ where $v$ and $p$ denote compensated velocity and reflected power, and a text prompt sequence $T=\{w_1, w_2, \dotsc, w_i\}$ with a length of $N_T$. The output of REC is a set of bounding boxes. The $i$-th bounding box is formulated as $B_i=\{x_i, y_i, w_i, h_i, c_{i}\}$, which represents the center coordinates, width, height and IoU confidence of bounding box. RES is to predict a mask of referred targets $M_{RES} \in R^{2 \times H_I \times H_w}$, where the channel dimension contains referred foreground target and unreferred background.

\subsection{Overall Pipeline}
\label{subsec:overall_pipeline}
\textbf{Feature Encoders.} \textbf{Firstly}, the image encoder adopts FastViT-SA24 \cite{vasu2023fastvit} with the cascade of reparametrized RepMixer and Multi-Head Self-Attention. \textbf{Secondly}, to prepare for fusion with image, we project 3D radar point clouds from radar to image coordinate, and obtain a 3-channel RVP map, where each position contains the target Range (R), compensated Velocity (V) and reflected Power (P). We adopt the combination of convolution, batch normalization and ReLU activation as the basic block to exact radar features. Each stage of radar encoder contains three blocks, where the convolution kernels are $3 \times 3$, $1 \times 1$ and $3 \times 3$ with the stride of $1$, $1$ and $2$ for local feature extraction, feedforward and downsampling. Whatever image or radar encoder, there are one stem layer and four stages, and the stem layer downsamples the spatial size four times. The feature dimension of stage $i \in \{1,2,3,4\}$ is $(64\times 2^{i-1}) \times \frac{I_H}{2^{i+1}} \times \frac{I_W}{2^{i+1}}$. 

\textbf{Feature Fusion.} To ensure precise guidance of prompt for both sensors to locate referred targets, the fusion with text requires high-quality integration of features from both sensors. Hence, we adopt a divide-and-conquer strategy called Phased Heterogeneous Modality Fusion. \textbf{Firstly}, due to water-surface reflection and multi-path effect from buildings on the riverbank, radar features unavoidably contain clutter in some scenarios, which results in a certain level of interference in fused sensor features. \textbf{Secondly}, not all prompts contain radar features of targets, so adaptive channel weighting matters. To tackle above problems, a lightweight module called Adaptive Radar Weighting is devised for clutter alleviation, where the last three stage features from the radar encoder will first go through three ARW modules before fusing with image features. Then, three-stage sensor features act as queries to fuse with the keys and values of textual features from the pretrained ALBERT \cite{lan2020albert} by our proposed low-power MHSCA and output three fusion features, which mitigates the quadratic complexity and high power consumption caused by vanilla MHCA \cite{wu2023referring}. A lite Path Aggregation Network (PANet) \cite{li2022yolov6} is to fuse features with various receptive fields, and outputs four-stage pyramid features from $P2$ to $P5$. 

\textbf{Prediction Heads.} We feed all four pyramid features to REC heads, which are adapted upon YOLOv8 decouple head \cite{Jocher_Ultralytics_YOLO_2023}. Each pyramid feature corresponds to one REC head, containing confidence and bounding box regression branch. The shape of REC output for the stage $i \in \{2,3,4,5\}$ is $(1+4\times RegMax) \times \frac{I_H}{2^{i+1}} \times \frac{I_W}{2^{i+1}}$, where $1$ is the confidence score and $4$ is four coordinates of bounding box. $RegMax$ is the hyperparameter of Distributed Focal Loss (DFL) \cite{li2020generalized}. Further, following the paradigm of PP-LiteSeg \cite{peng2022pp}, we feed $P2$, $P3$ and $P4$ with SPPM \cite{Jocher_Ultralytics_YOLO_2023} to the RES head, which is based on FLD \cite{peng2022pp}. The output shape of RES head is $2 \times I_H \times I_W$, including referred foreground and unreferred background categories.

\subsection{Phased Heterogeneous Modality Fusion}
\label{subsec:phmf}

\begin{figure}
    \includegraphics[width=0.99\linewidth]{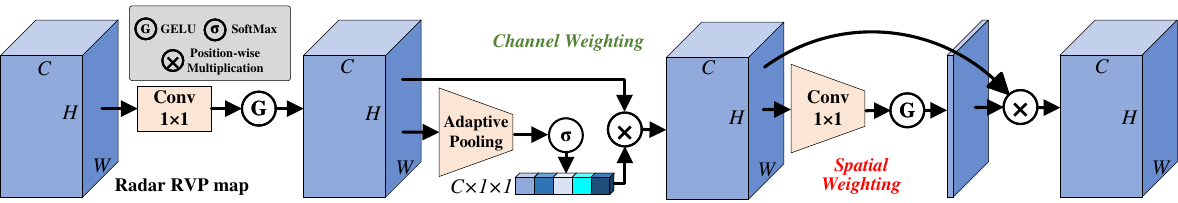}
    \vspace{-3mm}
    \caption{\textmd{The structure of Adaptive Radar Weighting.}}
    \label{fig:ARW}
\end{figure}

\begin{figure}
    \includegraphics[width=0.99\linewidth]{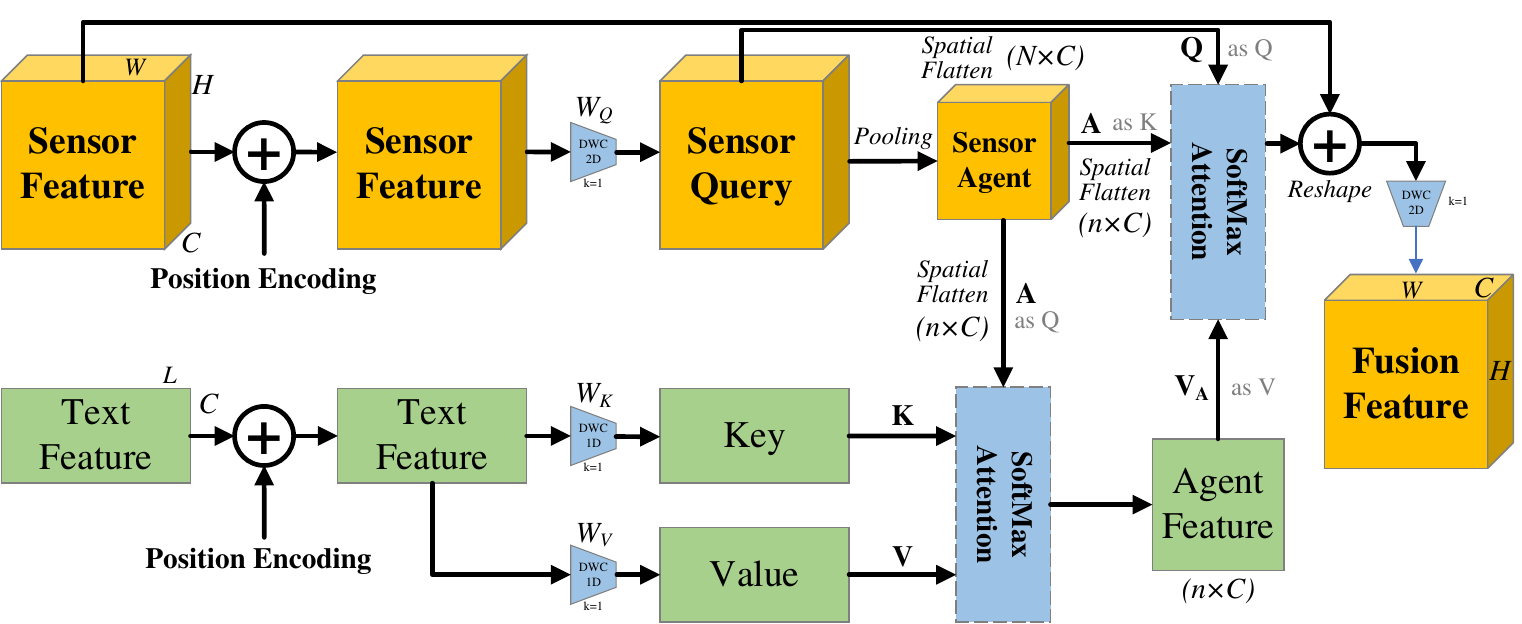}
    \vspace{-3mm}
    \caption{\textmd{The structure of Slim Cross Attention.}}
    \label{fig:sca}
\end{figure}

\textbf{(a) Adaptive Radar Weighting.} ARW is employed to dynamically weigh RVP maps on both channels and space to obtain high-quality features, where channel weighting is for aligning the text prompt containing corresponding radar features while spatial weighting is to alleviate the negative impact of radar clutter.

As Figure \ref{fig:ARW} presents, assuming the radar feature is $F_R \in R^{C \times H \times W}$, we exert a linear projection with GeLU activation on it to obtain $\hat{F_R} \in R^{C \times H \times W}$. The resulting feature map is then prepared for adaptive channel attention matrix $W_{CA} \in R^{C \times 1 \times 1}$ by applying adaptive average pooling to collapse spatial dimensions and computing attention weights by softmax. $\hat{F_R}$ is position-wise multiplication with $W_{CA}$ to obtain the channel-weighting feature $F_R^{CW} \in R^{C \times H \times W}$. Then spatial weighting matrix $W_{SA} \in R^{1 \times H \times W}$ is achieved by convolving $F_R^{CW}$ with a GELU activation and a 1 $\times$ 1 convolution, followed by element-wise multiplication with the original features and obtain the final radar feature $F_R^{ARW} \in R^{C \times H \times W}$ with both channel and spatial adaptive weighting. Equation \ref{eq:arw1} and \ref{eq:arw2} present the whole process above. 

\begin{align}
\left\{
    \begin{aligned}
    & \hat{F_R} = GeLU(Conv_{1\times 1}(F_R)), \hat{F_R} \in R^{C \times H \times W},
    \\
    & W_{CA} = SoftMax(AdaPool(\hat{F_R})), \hat{W_{CA}} \in R^{C \times 1 \times 1},
    \\
    & F_R^{CW} = \hat{F_R} \cdot W_{CA}, F_R^{CW} \in R^{C \times H \times W},
    \label{eq:arw1} \\
    \end{aligned}
    \right.
    \\
    \left\{
    \begin{aligned}
        & W_{SA} = GeLU(Conv_{1\times 1}(F_R^{CW})), W_{SA} \in R^{1 \times H \times W},
        \\
        & F_R^{ARW} = F_R^{CW} \cdot W_{SA}, F_R^{ARW} \in R^{C \times H \times W}.
        \label{eq:arw2}
    \end{aligned}
\right.
\end{align}


\textbf{(b) Multi-Head Slim Cross Attention.} After obtaining three-stage sensor features ($F^2_S$, $F^3_S$ and $F^4_S$) containing contextual features of vision and prior target features captured by radar, there are three linear projection branches from ALBERT to align the channel dimensions of three-stage sensor features. To effectively fuse two heterogeneous modalities with a low cost on USV-based edge devices, we propose Multi-Head Slim Cross Attention (MHSCA) (Figure \ref{fig:sca}). Given a sensor feature $F_S \in R^{C \times H \times W}$ and a text feature $F_T \in R^{L \times C}$, we first exert learnable encodings \cite{dosovitskiy2020image}\cite{vaswani2017attention}, $P_I$ and $P_T$ on them, respectively. Then we adopt 2D depth-wise convolution with $1 \times 1$ kernel ($W_Q$) to obtain the sensor query ($Q$) while two 1D depth-wise convolutions ($W_K$ and $W_V$) are for generate key ($K$) and value ($V$). Equation \ref{eq:pos_encoding} presents the aforementioned process.

\begin{equation}
    \begin{aligned}
    & Q = F_S W_Q + P_S, K = F_T W_K + P_T, V = F_V W_V + P_T, \\
    & Q \in R^{C \times H \times W}, K \in R^{L \times C}, V \in R^{L \times C},
    \label{eq:pos_encoding}
\end{aligned}
\end{equation}

Further, we employ adaptive pooling to downsample the sensor query $Q$ \cite{han2023agent}, preserving its essential context while reducing its spatial size to obtain the sensor agent $A \in R^{C \times h \times w}$. As the agent of $Q$, $A$ is firstly flattened along spatial dimension as $\hat{A} \in R^{n \times C}$ with $n=h\times w$, and conducts softmax attention $\sigma(\cdot)$ with $K$ and $V$ to calculate the agent-conditional Value matrix $V_A$. (Equation \ref{eq:agent_attn0}, \ref{eq:agent_attn1})

\begin{align}
    & \hat{A} = Flat(A)^T, \hat{A} \in R^{n \times C}, 
    \label{eq:agent_attn0} \\
    & V_A = \sigma(\hat{A}, K, V) = \frac{\hat{A}K^T}{\sqrt{d}}\cdot V,
    \label{eq:agent_attn1}
\end{align}
where $d$ is the channel dimension of $F_S$ and $F_T$.

Then, $V_A$ is forwarded to the second softmax attention, broadcasting the inter-modal similarity embedded in text feature to the sensor query $Q$, where $A$ acts as the key and $V_A$ is the value (Equation \ref{eq:agent_attn22}, \ref{eq:agent_attn2}). Here $Q$ is also firstly flattened as $\hat{Q} \in R^{N \times C}$ with $N=H \times W$.

\begin{align}
    & \hat{Q} = Flat(Q)^T, \hat{Q} \in R^{N \times C}, \label{eq:agent_attn22} \\
    & F_{TS} = \sigma(\hat{Q}, A, V_A) = \frac{\hat{Q}A^T}{\sqrt{d}}\cdot V_A,
    \label{eq:agent_attn2}
\end{align}

After that, a residual path from the sensor feature $F_S$ is added to the output of the second softmax attention and obtains the image-like sensor-conditional text features $F_{TS}$ (Equation \ref{eq:agent_attn3}). Finally, a 2D depth-wise convolution with $1 \times 1$ kernel size ($W$) is exerted on $F_{TS}$ to further feedforward.

\begin{equation}
    \hat{F}_{TS} = (Reshape(F_{TS}) + F_S)W, F_{TS} \in R^{C \times H \times W},
    \label{eq:agent_attn3}
\end{equation}

Here, the computation complexity of our proposed slim cross attention is $O(Nnd) + O(NLd)$, which is lower than vanilla cross attention's \cite{wu2023referring} $O(NLd) + O(NLd)$, as $n \ll L$ in practice.

\subsection{Training Objectives}
\label{subsec:training_obj}
We reformulate the REC and RES tasks as the bounding box regression and semantic segmentation. Though all targets are denoted as one category, referred target, implicit unbalance between target categories still exist. Hence, as Equation \ref{eq:rec_loss} presents, we introduce Distributed Focal Loss (DFL) \cite{li2020generalized} to alleviate the unbalanced distribution while CIoU \cite{zheng2020distance} is added to refinedly regress the bounding box size by taking its aspect ratio into account. 

\begin{equation}
    L_{REC} = L_{DFL} + L_{CIoU},
    \label{eq:rec_loss}
\end{equation}

For RES task, we combine the dice loss \cite{sudre2017generalised} with focal loss \cite{lin2017focal} to jointly subtly optimize the mask prediction (Equation \ref{eq:res_loss}).

\begin{equation}
    L_{RES} = L_{Dice} + L_{FL},
    \label{eq:res_loss}
\end{equation}

We empirically argue that the loss functions are not in the same order of magnitude for REC and RES during multi-task training. To avoid the domination of optimization by the task with a large loss, we introduce uncertainty weighting \cite{kendall2018multi} to balance multi-task optimization. Equation \ref{eq:joint_loss} demonstrates the loss function during the joint optimization of REC and RES.

\begin{equation}
    L_{Joint} = \frac{1}{2\sigma_1^2} L_{REC} + \frac{1}{2\sigma_2^2} L_{RES} + log\sigma_1 + log\sigma_2,
    \label{eq:joint_loss}
\end{equation}
where $\sigma_1$ and $\sigma_2$ are learnable parameters to weigh two loss items.

\section{Experiments}
\label{sec:experiment}
\subsection{Settings of Dataset and Implementation}

\textbf{Datasets.} Our proposed WaterVG is for training, validation and testing on all models. WaterVG includes two benchmarks, Fusion-based Visual Grounding (FVG) and Vision-based Visual Grounding (VVG). Moreover, to comprehensively valid the performance of Potamoi, we also evaluate our Potamoi on RefCOCO (testA, testB), RefCOCO+ (testA, testB) and RefCOCOg (test-umd) \cite{yu2016modeling}.

\textbf{Models.} For the REC, MattNet \cite{yu2018mattnet}, MCN \cite{luo2020multi}, TransVG \cite{deng2021transvg}, SimREC \cite{luo2022what}, SeqTR \cite{zhu2022seqtr}, LiteREC \cite{runwei2023literec}, RefTrans\cite{li2021referring} and our proposed Potamoi are included. For the RES, we introduce MCN, SeqTR, VLT \cite{ding2022vlt}, RefTrans and our Potamoi for evaluation. Furthermore, to equally evaluate these models on FVG, we add the same radar encoder and ARW as Potamoi for image and radar to each model. For Potamoi, we adopt ALBERT tokenizer \cite{lan2020albert} with a maximum token length of 50. Besides, the pretrained weights of FastViT-SA24 (image encoder) and other models' backbones on ImageNet-1K \cite{deng2009imagenet} are loaded during training. The head number of proposed MHSCA are 4, 4 and 8 for three stages while the length of sensor agent $n$ is set as 49. On RefCOCOs, our Potamoi is trained from scratch.

\textbf{Training.} We resize the input image and RVP map as 640 $\times$ 640 (px). We train all models for 80 epochs on WaterVG with a batch size of 16 while 100 epochs with 64 batch size on RefCOCOs by mixed-precision mode. We set the initial learning rate as 1e-3 with the cosine scheduler. We adopt SGDM as the optimizer with a momentum of 0.937 and a weight decay of 5e-4. All training experiments are on four RTX A4000 GPUs.

\textbf{Evaluation metrics.} Compared with datasets \cite{yu2016modeling,plummer2015flickr30k} where a prompt refers to one target, our WaterVG involves a single text prompt referring to multi-targets. Hence, to comprehensively evaluate model performances, we use $AP_{50}$ (Average Precision above IoU 0.5), $AP_{50-95}$ and $AR_{50-95}$ (Average Precision and Recall from IoU 0.5 to 0.95) in REC. For RES tasks, we adopt mIoU to evaluate models. Moreover, we evaluate models on their power consumption \cite{tu2023femtodet}. In tables for comparison, bold text indicates the best, and underlined text indicates the second best. For RefCOCOs, we use $AP_{50}$ as the metric of REC while mIoU as the RES metric.

\begin{figure}
    \includegraphics[width=1.00\linewidth]{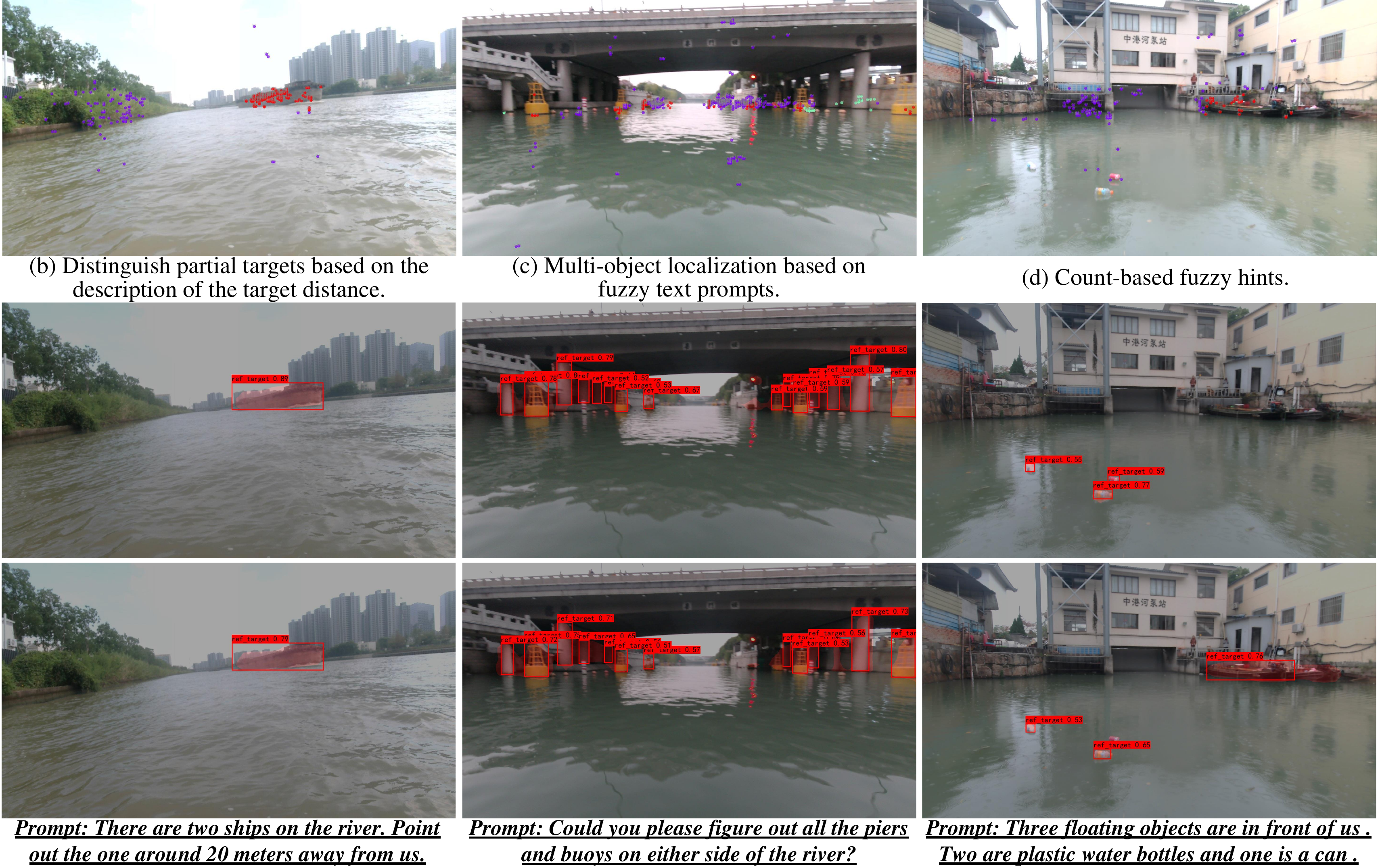}
    \vspace{-5mm}
    \caption{\textmd{Images with projected radar point clouds (PCs) (first row) and prediction results by Potamoi (second row) and MCN \cite{luo2020multi} (third row) under different situations. Red bounding boxes are REC predictions while translucent red masks are RES predictions. Blue PCs denote clutter while others represent targets. The gray letters on PCs refers to the distance (m) from targets to the ego USV.}}
    \label{fig:pred_potamoi}
\end{figure}

\begin{table*}
\caption{\textmd{Overall comparison of models on WaterVG. \textbf{MT} means multi-task while $P$ is power consumption calculated by hardware API.}}
\normalsize
\vspace{-3mm}
    \label{tab:overall_compare}
    \setlength\tabcolsep{0.25pt}
    \centering
    \begin{tabular}{c|c|ccc|c|ccc|c|c|ccc|c|ccc|c|c}
    \toprule
    & & \multicolumn{8}{c}{\textbf{Vision-based Visual Grounding (VVG)}} & \multicolumn{9}{c}{\textbf{Fusion-based Visual Grounding (FVG)}} \\
    \midrule
    \multirow{2}[2]{*}{\textbf{Methods}} & \multirow{2}[2]{*}{\textbf{MT}} &  \multicolumn{4}{c}{\textbf{Val}} & \multicolumn{4}{c}{\textbf{Test}} & & \multicolumn{4}{c}{\textbf{Val}} & \multicolumn{4}{c}{\textbf{Test}} &  \\
    \cmidrule(lr){3-6} \cmidrule(lr){7-10} \cmidrule(lr){12-15} \cmidrule(lr){16-19}
     & & \textbf{AP$_{50}$} & \textbf{AP$_{50\text{-}95}$} & \textbf{AR$_{50\text{-}95}$} &  \textbf{mIoU} & \textbf{AP$_{50}$} & \textbf{AP$_{50\text{-}95}$} & \textbf{AR$_{50\text{-}95}$} & \textbf{mIoU} & \textbf{P} & \textbf{AP$_{50}$} & \textbf{AP$_{50\text{-}95}$} & \textbf{AR$_{50\text{-}95}$} &  \textbf{mIoU} & \textbf{AP$_{50}$} & \textbf{AP$_{50\text{-}95}$} & \textbf{AR$_{50\text{-}95}$} &  \textbf{mIoU} & \textbf{P}\\
    \midrule
      MattNet & \Checkmark & 59.6 & 31.2 & 39.7 & 54.81 & 57.2 & 30.2 & 38.9 & 54.72 & - & 63.9 & 38.1 & 44.0 & 57.03 & 59.1 & 32.3 & 38.9 & 60.32 & - \\
      MCN & \Checkmark & 69.9 & 39.3 & 50.7 & 60.16 & 66.7 & 39.4 & 46.2 & 59.05 & 81.34 & 70.6 & 45.6  & 52.5 & 61.72 & 68.1 & 42.5 & 46.9 & 66.23 &  77.72\\
      TransVG & \XSolidBrush & 70.1 & 40.7 & 51.0 & - & 67.2 & 39.8 & 48.0 & - & - & 71.8 & 47.1 & 54.0 & - & 69.8 & \uline{43.6} & 52.0 & - & - \\
      SimREC & \XSolidBrush & 56.7 & 31.3 & 39.8 & - & 54.0 & 30.1 & 37.9 & - & 94.42 & 60.5 & 36.8 &  42.5 & - & 56.1 & 30.9 & 39.7 & - & 88.70 \\
      SeqTR & \Checkmark & 71.2 & 42.0 & 51.8 & 62.98 & 69.1 & 41.7 & 49.2 & 60.03 & 90.61 & 72.6 & \textbf{47.7} & \uline{54.2} & 64.96 & \uline{70.0} & 43.5 & 52.6 & 68.79 & 84.65 \\
      LiteREC & \XSolidBrush & 65.5 & 37.9 & 44.9 & - & 63.5 & 36.3 & 46.3 & - & \uline{70.22} & 68.9  & 42.7 & 47.6 & -  & 64.2 & 36.3 & 44.5 & - & \uline{65.79} \\
      RefTrans &  \Checkmark & 70.8 & 42.8 & 52.9 & \uline{63.72} & \textbf{70.8} & \uline{42.5} & \uline{50.1} & \uline{61.03} & 92.97 & \textbf{73.1} & 47.1 & 53.8 & 66.19 & 69.7 & 43.1 & \uline{52.7} & 70.19 & 87.21 \\
      VLT  & \XSolidBrush & - & - & - & \textbf{64.13} &  - & - & - & \textbf{61.79} & - & - & - & - & \textbf{67.59} & - & - & - & \textbf{70.90} & - \\
      \midrule
      \textbf{Potamoi} & \Checkmark & \textbf{71.6} & \textbf{43.3} & \textbf{54.9} & 63.16 & \uline{70.0} & \textbf{42.6} & \textbf{50.7} & 60.36 & \textbf{66.10} & \uline{72.8} & \uline{47.5} & \textbf{55.0} & \uline{66.27} & \textbf{70.1} & \textbf{44.8} & \textbf{53.0} & \uline{70.81} & \textbf{61.48} \\
    \bottomrule
    \end{tabular}
\end{table*}

\begin{table}[h]
    \caption{\textmd{Potamoi under waterways, sentence and query patterns.}}
    \vspace{-3mm}
    \label{tab:propotion_experiments}
    \setlength\tabcolsep{3.6pt}
    \centering
    \begin{tabular}{c|cccc}
    \toprule
     \textbf{Waterways} & \textbf{Canal} & \textbf{Moat} & \textbf{River} & \textbf{Lake} \\
     \midrule
       \textbf{AP$_{50}$}  & 72.5 & 69.4 & 70.5 & 67.5 \\
       \textbf{mIoU} & 72.90 & 71.81 & 73.45 & 66.48 \\
     \midrule
     \textbf{Sentences} & \textbf{SE$^1$} & \textbf{SI$^2$} & \textbf{QU$^3$} & \textbf{EX$^4$} \\
     \midrule
     \textbf{AP$_{50}$} & 73.6 & 70.0 & 65.5 & 70.1 \\
     \textbf{mIoU} & 74.12 & 71.53 & 66.41 & 68.74 \\
     \midrule
     \textbf{Queries} & \textbf{Exact-Feat} & \textbf{Partial-Feat} & \textbf{Number} & \textbf{Category} \\ 
     \midrule
     \textbf{AP$_{50}$} & 73.1 & 70.1 & 62.2 & 68.7 \\
     \textbf{mIoU} & 74.35 & 71.76 & 63.57 & 66.87 \\
     \bottomrule
    \end{tabular}
    \footnotesize{1. Statement (egocentric), 2. Statement (imperative), 3. Question, 4. Exclamation.}
\end{table}

\subsection{Quantitative Results}

Table \ref{tab:overall_compare} presents the overall performances of our proposed Potamoi and other competitors on the benchmarks of VVG and FVG. \textbf{Firstly}, overall, we find that all models perform better on FVG than on VVG. This suggests that relying solely on visual features may not accurately localize specific described targets, particularly those incorporating distance or motion features in textual descriptions. \textbf{Secondly}, our proposed Potamoi achieves competitive performance compared to other visual grounding models, whether based on transformer architecture (TransVG, SeqTR, RefTrans) or one-stage (MCN, SimREC, LiteREC-S2) or two-stage paradigms (MattNet). \textbf{Specifically}, for REC task of VVG and FVG, our Potamoi achieves state-of-the-art performances overall, which in a race with RefTrans \cite{li2021referring} and SeqTR \cite{zhu2022seqtr}. \textbf{Moreover}, for RES task, due to the lite segmentation head \cite{peng2022pp} in Potamoi for power saving, it generally ranks second or third among all in Table \ref{tab:overall_compare}, trailing behind the dedicated RES model VLT \cite{ding2022vlt} based on transformer architecture with a larger parameter count. Nevertheless, among multi-task models, our Potamoi overperforms MCN, SeqTR and is evenly contested to RefTrans. Table \ref{tab:propotion_experiments} presents the performance of Potamoi on test set under different scenarios and sentence query patterns as Figure \ref{fig:dataset_statistics_1} presents. \textbf{Importantly}, our Potamoi achieves the lowest power consumption for both VVG or FVG, which is much lower than both single-task and multi-task models, implying Potamoi can help extend the endurance time of USV operations. \textbf{Lastly}, Table \ref{tab:compare_rec} and \ref{tab:compare_res} show that our low-power Potamoi obtains competitive performances compared with other state-of-the-art models on RefCOCOs. Further, our MHSCA can be nicely adapted to CNN-based MCN and transformer-based SeqTR to improve performance.

\begin{table}
    \caption{\textmd{Text-sensor fusion methods in Potamoi on the test set.}}
    \vspace{-3mm}
    \label{tab:attncompare}
    \setlength\tabcolsep{2.7pt}
    \centering
    \small
    \begin{tabular}{c|cc|cccc} 
    \toprule
        \textbf{Methods} & \textbf{Params}$\downarrow$ & \textbf{FLOPs}$\downarrow$ & \textbf{AP$_{50}$}$\uparrow$ & \textbf{AP$_{50\text{-}95}$}$\uparrow$ & \textbf{AR$_{50\text{-}95}$}$\uparrow$ &  \textbf{mIoU}$\uparrow$ \\
    \midrule
    \multicolumn{6}{c}{\textcolor{green}{Stage 2: }\textbf{$D$=128, $N_h$=4, $H$=80, $W$=80, $L$=50}} \\
    \multicolumn{6}{c}{\textcolor{blue}{Stage 3: }\textbf{$D$=256, $N_h$=4, $H$=40, $W$=40, $L$=50}} \\
    \multicolumn{6}{c}{\textcolor{red}{Stage 4: }\textbf{$D$=512, $N_h$=8, $H$=20, $W$=20, $L$=50}} \\
    \midrule
    \multirow{3}[2]{*}{MHCA \cite{dao2022flashattention}} & \textcolor{green}{\uline{66.048K}} & \textcolor{green}{293.27M} & \multirow{3}[2]{*}{\uline{71.8}} & \multirow{3}[2]{*}{45.0} & \multirow{3}[2]{*}{52.9} & \multirow{3}[2]{*}{70.23} \\
                                                          & \textcolor{blue}{\uline{0.263M}} & \textcolor{blue}{257.23M}   \\
                                                         & \textcolor{red}{1.050M}  & \textcolor{red}{256.41M}   \\
    \midrule
    \multirow{3}[2]{*}{MHLCA \cite{choromanski2020rethinking}} & \textcolor{green}{0.252M} & \textcolor{green}{1.691G}  &  \multirow{3}[2]{*}{70.2} & \multirow{3}[2]{*}{44.7} & \multirow{3}[2]{*}{\uline{53.2}} & \multirow{3}[2]{*}{\textbf{72.01}} \\
                                           & \textcolor{blue}{0.524M} & \textcolor{blue}{865.08M}  \\
                                          & \textcolor{red}{\uline{1.049M}} & \textcolor{red}{471.86M} \\
    \midrule
    \multicolumn{6}{c}{\textbf{Default Length of Sensor Agent $n$ = 49 (7 $\times$ 7)}} \\
    \midrule
    \multirow{3}[2]{*}{\textbf{MHSCA}} & \textcolor{green}{\textbf{0.768K}} & \textcolor{green}{\textbf{8.11M}} & \multirow{3}[2]{*}{70.1} & \multirow{3}[2]{*}{44.8} & \multirow{3}[2]{*}{53.0} & \multirow{3}[2]{*}{70.81} \\
                                         & \textcolor{blue}{\textbf{1.536K}} & \textcolor{blue}{\textbf{3.98M}} \\
                                         & \textcolor{red}{\textbf{3.072K}} & \textcolor{red}{\textbf{1.99M}} \\
    \midrule
    \multicolumn{6}{c}{\textbf{Length of Sensor Agent $n$ = 144 (12 $\times$ 12)}} \\
    \midrule
    \multirow{3}[2]{*}{\textbf{MHSCA}} & \textcolor{green}{\textbf{0.768K}} & \textcolor{green}{\uline{8.14M}} & \multirow{3}[2]{*}{71.7} & \multirow{3}[2]{*}{\uline{45.1}} & \multirow{3}[2]{*}{53.1} & \multirow{3}[2]{*}{71.09}\\
                                       & \textcolor{blue}{\textbf{1.536K}} & \textcolor{blue}{\uline{4.07M}}  \\
                                       & \textcolor{red}{\textbf{3.072K}} & \textcolor{red}{\uline{2.04M}}  \\
    \midrule
    \multicolumn{6}{c}{\textbf{Length of Sensor Agent $n$ = 256 (16 $\times$ 16)}} \\
    \midrule
    \multirow{3}[2]{*}{\textbf{MHSCA}} & \textcolor{green}{\textbf{0.768K}} & \textcolor{green}{8.24M} & \multirow{3}[2]{*}{\textbf{72.3}} & \multirow{3}[2]{*}{\textbf{45.2}} & \multirow{3}[2]{*}{\textbf{54.0}} & \multirow{3}[2]{*}{\uline{71.95}}\\
                                       & \textcolor{blue}{\textbf{1.536K}} & \textcolor{blue}{\textbf{3.98M}} \\
                                       & \textcolor{red}{\textbf{3.072K}} & \textcolor{red}{2.26M}  \\
    \bottomrule
    \end{tabular}
\end{table}

\begin{figure*}
    \includegraphics[width=1.00\linewidth]{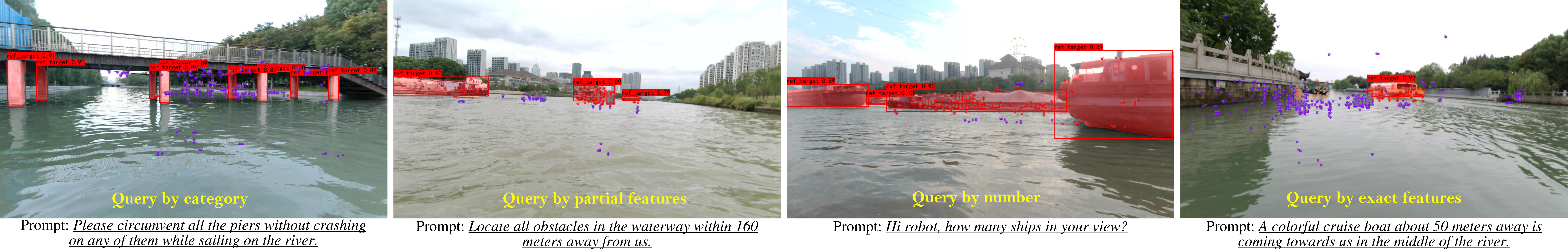}
    \vspace{-5mm}
    \caption{\textmd{Prediction samples of Potamoi on four primary query types of textual prompts.}}
    \label{fig:query_pred}
\end{figure*}

\subsection{Qualitative Results}
Figure \ref{fig:pred_potamoi} visualizes predictions of our Potamoi and MCN \cite{luo2020multi} on the representative samples of WaterVG, including (a) distinguish partial targets based on the query of target distance, (b) multi-target localization based on fuzzy text prompts and (c) count-based fuzzy hints. Apparently, our Potamoi behaves more prominent than MCN, obtaining lower target miss rates, false positive rates, and masks that better fit shapes and contours of the targets. Moreover, Figure \ref{fig:query_pred} presents prediction samples upon four primary query types in WaterVG, where our Potamoi exhibits nice adaptation.

\subsection{Ablation Experiments}

\textbf{Cross attention comparison.} In Table \ref{tab:attncompare}, we conduct a comprehensive comparison between our MHSCA and two representative multi-head cross-attention mechanisms, MHCA \cite{dao2022flashattention} and MHLCA \cite{choromanski2020rethinking}, which are transplanted to our Potamoi, encompassing comparisons across different channel dimensions and spatial sizes of visual feature maps in PAN. Moreover, we adjust the agent number $n$ in MHSCA as 49, 144 and 256 for comparison, respectively. 

\begin{figure}
    \includegraphics[width=0.99\linewidth]{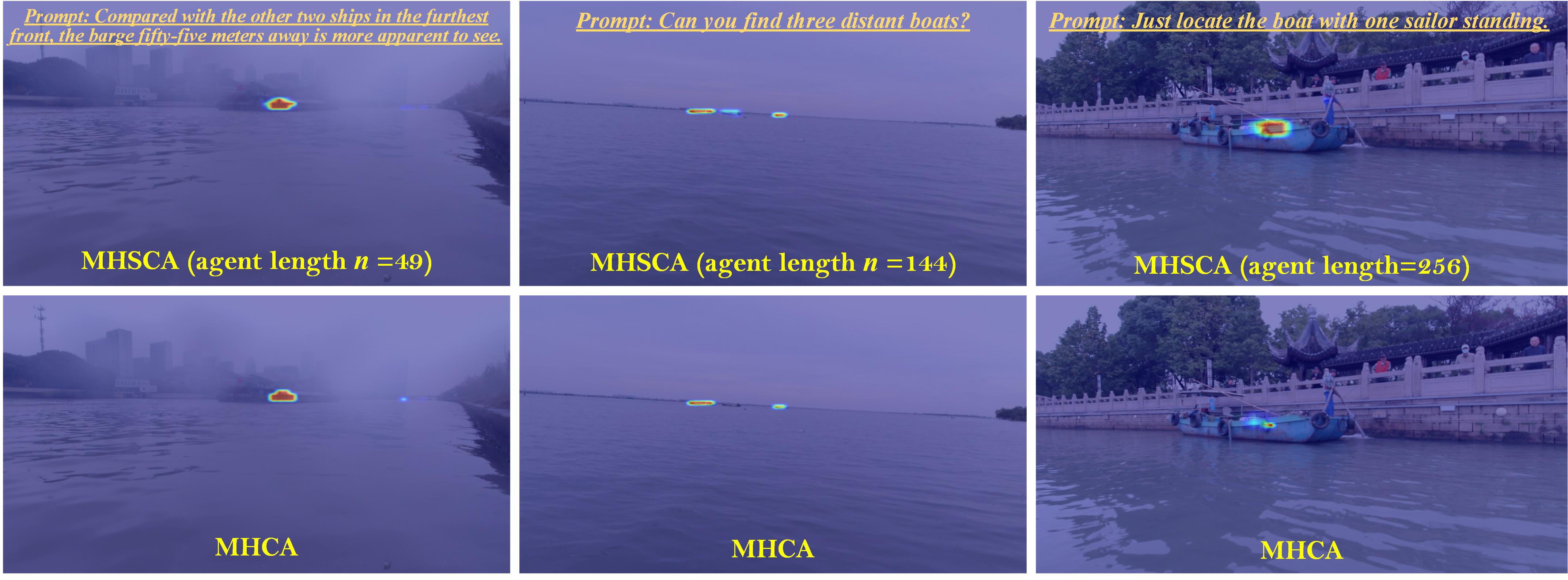}
    \vspace{-3mm}
    \caption{\textmd{Last-stage heatmaps of PAN before prediction heads.}}
    \label{fig:heatmaps}
\end{figure}

Firstly, looking at the performance on both REC and RES tasks, our proposed MHSCA ($n$=256) achieves the overall best results. MHSCA ($n$=144) also obtains competitive performances. It is noteworthy that for visual feature maps with increasing spatial sizes, the FLOPs of our MHSCA never exceed 10M and demonstrate smaller growth compared to flash attention-based \cite{dao2022flashattention} MHCA and MHLCA. Regarding expanding channels, our MHSCA maintains a significantly lower parameter count compared to the other two fusion methods. Further, Figure \ref{fig:heatmaps} visualizes the last stage of PAN after fusing multi-scale attention maps under settings of MHSCA and MHCA, MHSCA correctly capture targets referred by prompt, also advanced in small target localization. Therefore, MHSCA achieves a well-balanced result between accuracy and complexity, serving as a lite and scalable cross-attention method for multi-scale maps.

\begin{table}
    \caption{\textmd{Comparison of REC on RefCOCOs datasets.}}
    \setlength\tabcolsep{3.7pt}
    \vspace{-3mm}
    \centering
    \begin{tabular}{c|cc|cc|c}
    \toprule
      \multirow{2}[2]{*}{\textbf{Models}} & \multicolumn{2}{c}{\textbf{RefCOCO}} & \multicolumn{2}{c}{\textbf{RefCOCO+}} & \textbf{RefCOCOg} \\
      \cmidrule(lr){2-3} \cmidrule(lr){4-5} \cmidrule(lr){6-6}
       & testA & testB & testA & testB & test-u \\
      \midrule
      MAttNet & 80.43 & 69.28 & 70.26 & 56.00 & 67.27 \\
      TransVG & 82.72 & \textbf{78.35} & 70.70 & 56.94 & 67.73 \\
      RefTrans & \textbf{85.53} & 76.31 & 75.58 & \textbf{61.91} & 69.10 \\
      \midrule
      MCN & 82.29 & 74.98 & 72.86 & 57.31 & 66.01 \\
      MCN-\textbf{MHSCA} & 83.06$\uparrow$ & 75.22$\uparrow$ & 73.54$\uparrow$ & 58.67$\uparrow$ & 66.92$\uparrow$ \\
      SeqTR & 85.00 & 76.08 & 75.37 & 58.78 & \uline{71.58} \\
      SeqTR-\textbf{MHSCA} & \uline{85.36}$\uparrow$ & 76.72$\uparrow$ & \uline{75.85}$\uparrow$ & \uline{59.10}$\uparrow$ & \textbf{72.13}$\uparrow$ \\
      \midrule
      \textbf{Potamoi} & 84.98 & \uline{78.10} & \textbf{75.90} & 58.55 & 70.55 \\
    \bottomrule
    \end{tabular}
    \label{tab:compare_rec}
\end{table}

\begin{table}
    \caption{\textmd{Comparison of RES on RefCOCOs datasets.}}
    \setlength\tabcolsep{3.3pt}
    \vspace{-3mm}
    \centering
    \begin{tabular}{c|cc|cc|c}
    \toprule
      \multirow{2}[2]{*}{\textbf{Models}} & \multicolumn{2}{c}{\textbf{RefCOCO}} & \multicolumn{2}{c}{\textbf{RefCOCO+}} & \textbf{RefCOCOg} \\
      \cmidrule(lr){2-3} \cmidrule(lr){4-5} \cmidrule(lr){6-6}
       & testA & testB & testA & testB & test-u \\
      \midrule
      MAttNet & 62.37 & 51.70 & 52.39 & 40.08 & 48.61 \\\
      VLT & 68.29 & 62.73 & \textbf{59.20} & \textbf{49.36} & \textbf{56.65} \\
      \midrule
      MCN & 64.20 & 59.71 & 54.99 & 44.69 & 49.40 \\
      MCN-\textbf{MHSCA} & 64.51$\uparrow$ & 59.94$\uparrow$ & 55.30$\uparrow$ & 45.32$\uparrow$ & 49.76$\uparrow$ \\
      SeqTR & \uline{69.79} & \uline{64.12} & 58.93 & 48.19 & 55.64 \\
      SeqTR-\textbf{MHSCA} & \textbf{70.10}$\uparrow$ & \textbf{64.32}$\uparrow$ & \uline{59.09}$\uparrow$ & 48.33$\uparrow$ & \uline{55.90}$\uparrow$ \\
      \midrule
      \textbf{Potamoi} & 69.05 & 63.67 & 59.05 & \uline{48.72} & 55.23 \\
    \bottomrule
    \end{tabular}
    \label{tab:compare_res}
\end{table}

\textbf{Comparison of other settings.}  Table \ref{tab:ablation_other_settings} presents ablation experiments of other settings in fusion-based Potamoi, including settings of input data, network and training. For radar features, target range and velocity are more vital for visual grounding. Besides, due to some samples containing prompt token lengths exceeding 30, and certain target-descriptive terms occurring towards the end of text, reducing token length resulted in information loss, leading to a decrease in accuracy. Moreover, when removing ARW and only with element-wise addition for vision-radar fusion, an apparent drop appears. When adopting multi-task learning, the performance of Potamoi is better than single-task learning, which indicates multi-task learning can promote performances of both tasks.

\vspace{-3mm}
\begin{table}
    \caption{\textmd{Ablation experiments on other settings of Potamoi.}}
    \vspace{-3mm}
    \label{tab:ablation_other_settings}
    \setlength\tabcolsep{7.9pt}
    \centering
    \begin{tabular}{c|cccc}
    \toprule
     \textbf{Methods} & \textbf{AP$_{50}$} & \textbf{AP$_{50\text{-}95}$} & \textbf{AR$_{50\text{-}95}$} & \textbf{mIoU}  \\
     \midrule
     \textbf{Baseline} & \uline{70.1} & \textbf{44.8} & \textbf{53.0} & \textbf{70.81} \\
     \midrule
     \multicolumn{5}{c}{\textbf{Input Data}} \\
     \midrule
     Radar map (R) & 70.0 & 42.8 & \uline{52.5} & 70.41  \\ 
     Radar map (R-V) & \uline{70.1} & 43.1 & 52.4 & \uline{70.55} \\
     Radar map (R-P) & 69.8 & 42.9 & 52.4 & 70.24 \\
     Radar map (V-P) & 69.6 & 42.8 & 52.3 &  69.53 \\
     \midrule
     Token Len=30 & 69.8 & \uline{43.2} & 52.0 & 70.31 \\
     \midrule
     \multicolumn{5}{c}{\textbf{Network Settings}} \\
     \midrule
     No ARW & 69.5 & 42.8 & 52.2 & 70.40\\
     \midrule
     \multicolumn{5}{c}{\textbf{Training}} \\
     \midrule
     Single-task  & \textbf{70.9} & 44.0 & 52.6 & 70.68 \\
     \bottomrule
    \end{tabular}
\end{table}

\section{Conclusion}
\label{sec:conclusion}
We propose the first visual grounding dataset of USV-based waterway perception, WaterVG, which includes fine-grained textual prompts, visual, and millimeter-wave radar data. Further, each sample's text prompt contains references to multiple targets. Based on this, we design a multi-task visual grounding model, Potamoi, which can accurately performing REC and RES tasks simultaneously with the lowest power consumption among all models. This capability relies on our proposed ARW and highly efficient Multi-Head Slim Cross Attention (MHSCA) for fusion of textual and sensor features with a considerably low cost. Both Potamoi and MHSCA demonstrate competitive performance compared to counterparts.


\newpage
\normalem
\bibliographystyle{ACM_Reference_Format}
\bibliography{sample_sigconf}


\begin{thebibliography}{69}


\ifx \showCODEN    \undefined \def \showCODEN     #1{\unskip}     \fi
\ifx \showDOI      \undefined \def \showDOI       #1{#1}\fi
\ifx \showISBNx    \undefined \def \showISBNx     #1{\unskip}     \fi
\ifx \showISBNxiii \undefined \def \showISBNxiii  #1{\unskip}     \fi
\ifx \showISSN     \undefined \def \showISSN      #1{\unskip}     \fi
\ifx \showLCCN     \undefined \def \showLCCN      #1{\unskip}     \fi
\ifx \shownote     \undefined \def \shownote      #1{#1}          \fi
\ifx \showarticletitle \undefined \def \showarticletitle #1{#1}   \fi
\ifx \showURL      \undefined \def \showURL       {\relax}        \fi
\providecommand\bibfield[2]{#2}
\providecommand\bibinfo[2]{#2}
\providecommand\natexlab[1]{#1}
\providecommand\showeprint[2][]{arXiv:#2}

\bibitem[Bovcon et~al\mbox{.}(2019)]%
        {bovcon2019mastr1325}
\bibfield{author}{\bibinfo{person}{Borja Bovcon}, \bibinfo{person}{Jon Muhovi{\v{c}}}, \bibinfo{person}{Janez Per{\v{s}}}, {and} \bibinfo{person}{Matej Kristan}.} \bibinfo{year}{2019}\natexlab{}.
\newblock \showarticletitle{The mastr1325 dataset for training deep usv obstacle detection models}. In \bibinfo{booktitle}{\emph{2019 IEEE/RSJ International Conference on Intelligent Robots and Systems (IROS)}}. IEEE, \bibinfo{pages}{3431--3438}.
\newblock


\bibitem[Bovcon et~al\mbox{.}(2018)]%
        {bovcon2018stereo}
\bibfield{author}{\bibinfo{person}{Borja Bovcon}, \bibinfo{person}{Janez Per{\v{s}}}, \bibinfo{person}{Matej Kristan}, {et~al\mbox{.}}} \bibinfo{year}{2018}\natexlab{}.
\newblock \showarticletitle{Stereo obstacle detection for unmanned surface vehicles by IMU-assisted semantic segmentation}.
\newblock \bibinfo{journal}{\emph{Robotics and Autonomous Systems}}  \bibinfo{volume}{104} (\bibinfo{year}{2018}), \bibinfo{pages}{1--13}.
\newblock


\bibitem[Chen et~al\mbox{.}(2018)]%
        {chen2018real}
\bibfield{author}{\bibinfo{person}{Xinpeng Chen}, \bibinfo{person}{Lin Ma}, \bibinfo{person}{Jingyuan Chen}, \bibinfo{person}{Zequn Jie}, \bibinfo{person}{Wei Liu}, {and} \bibinfo{person}{Jiebo Luo}.} \bibinfo{year}{2018}\natexlab{}.
\newblock \showarticletitle{Real-time referring expression comprehension by single-stage grounding network}.
\newblock \bibinfo{journal}{\emph{arXiv preprint arXiv:1812.03426}} (\bibinfo{year}{2018}).
\newblock


\bibitem[Chen et~al\mbox{.}(2020)]%
        {chen2020cops}
\bibfield{author}{\bibinfo{person}{Zhenfang Chen}, \bibinfo{person}{Peng Wang}, \bibinfo{person}{Lin Ma}, \bibinfo{person}{Kwan-Yee~K Wong}, {and} \bibinfo{person}{Qi Wu}.} \bibinfo{year}{2020}\natexlab{}.
\newblock \showarticletitle{Cops-ref: A new dataset and task on compositional referring expression comprehension}. In \bibinfo{booktitle}{\emph{Proceedings of the IEEE/CVF Conference on Computer Vision and Pattern Recognition}}. \bibinfo{pages}{10086--10095}.
\newblock


\bibitem[Cheng et~al\mbox{.}(2021a)]%
        {cheng2021we}
\bibfield{author}{\bibinfo{person}{Yuwei Cheng}, \bibinfo{person}{Mengxin Jiang}, \bibinfo{person}{Jiannan Zhu}, {and} \bibinfo{person}{Yimin Liu}.} \bibinfo{year}{2021}\natexlab{a}.
\newblock \showarticletitle{Are we ready for unmanned surface vehicles in inland waterways? The usvinland multisensor dataset and benchmark}.
\newblock \bibinfo{journal}{\emph{IEEE Robotics and Automation Letters}} \bibinfo{volume}{6}, \bibinfo{number}{2} (\bibinfo{year}{2021}), \bibinfo{pages}{3964--3970}.
\newblock


\bibitem[Cheng et~al\mbox{.}(2021b)]%
        {cheng2021robust}
\bibfield{author}{\bibinfo{person}{Yuwei Cheng}, \bibinfo{person}{Hu Xu}, {and} \bibinfo{person}{Yimin Liu}.} \bibinfo{year}{2021}\natexlab{b}.
\newblock \showarticletitle{Robust small object detection on the water surface through fusion of camera and millimeter wave radar}. In \bibinfo{booktitle}{\emph{Proceedings of the IEEE/CVF International Conference on Computer Vision}}. \bibinfo{pages}{15263--15272}.
\newblock


\bibitem[Cheng et~al\mbox{.}(2021c)]%
        {cheng2021flow}
\bibfield{author}{\bibinfo{person}{Yuwei Cheng}, \bibinfo{person}{Jiannan Zhu}, \bibinfo{person}{Mengxin Jiang}, \bibinfo{person}{Jie Fu}, \bibinfo{person}{Changsong Pang}, \bibinfo{person}{Peidong Wang}, \bibinfo{person}{Kris Sankaran}, \bibinfo{person}{Olawale Onabola}, \bibinfo{person}{Yimin Liu}, \bibinfo{person}{Dianbo Liu}, {et~al\mbox{.}}} \bibinfo{year}{2021}\natexlab{c}.
\newblock \showarticletitle{Flow: A dataset and benchmark for floating waste detection in inland waters}. In \bibinfo{booktitle}{\emph{Proceedings of the IEEE/CVF International Conference on Computer Vision}}. \bibinfo{pages}{10953--10962}.
\newblock


\bibitem[Choromanski et~al\mbox{.}(2020)]%
        {choromanski2020rethinking}
\bibfield{author}{\bibinfo{person}{Krzysztof~Marcin Choromanski}, \bibinfo{person}{Valerii Likhosherstov}, \bibinfo{person}{David Dohan}, \bibinfo{person}{Xingyou Song}, \bibinfo{person}{Andreea Gane}, \bibinfo{person}{Tamas Sarlos}, \bibinfo{person}{Peter Hawkins}, \bibinfo{person}{Jared~Quincy Davis}, \bibinfo{person}{Afroz Mohiuddin}, \bibinfo{person}{Lukasz Kaiser}, {et~al\mbox{.}}} \bibinfo{year}{2020}\natexlab{}.
\newblock \showarticletitle{Rethinking Attention with Performers}. In \bibinfo{booktitle}{\emph{International Conference on Learning Representations}}.
\newblock


\bibitem[Chrupa{\l}a(2022)]%
        {chrupala2022visually}
\bibfield{author}{\bibinfo{person}{Grzegorz Chrupa{\l}a}.} \bibinfo{year}{2022}\natexlab{}.
\newblock \showarticletitle{Visually grounded models of spoken language: A survey of datasets, architectures and evaluation techniques}.
\newblock \bibinfo{journal}{\emph{Journal of Artificial Intelligence Research}}  \bibinfo{volume}{73} (\bibinfo{year}{2022}), \bibinfo{pages}{673--707}.
\newblock


\bibitem[Dao et~al\mbox{.}(2022)]%
        {dao2022flashattention}
\bibfield{author}{\bibinfo{person}{Tri Dao}, \bibinfo{person}{Daniel~Y. Fu}, \bibinfo{person}{Stefano Ermon}, \bibinfo{person}{Atri Rudra}, {and} \bibinfo{person}{Christopher R{\'e}}.} \bibinfo{year}{2022}\natexlab{}.
\newblock \showarticletitle{Flash{A}ttention: Fast and Memory-Efficient Exact Attention with {IO}-Awareness}. In \bibinfo{booktitle}{\emph{Advances in Neural Information Processing Systems}}.
\newblock


\bibitem[Deng et~al\mbox{.}(2009)]%
        {deng2009imagenet}
\bibfield{author}{\bibinfo{person}{Jia Deng}, \bibinfo{person}{Wei Dong}, \bibinfo{person}{Richard Socher}, \bibinfo{person}{Li-Jia Li}, \bibinfo{person}{Kai Li}, {and} \bibinfo{person}{Li Fei-Fei}.} \bibinfo{year}{2009}\natexlab{}.
\newblock \showarticletitle{Imagenet: A large-scale hierarchical image database}. In \bibinfo{booktitle}{\emph{2009 IEEE conference on computer vision and pattern recognition}}. Ieee, \bibinfo{pages}{248--255}.
\newblock


\bibitem[Deng et~al\mbox{.}(2021)]%
        {deng2021transvg}
\bibfield{author}{\bibinfo{person}{Jiajun Deng}, \bibinfo{person}{Zhengyuan Yang}, \bibinfo{person}{Tianlang Chen}, \bibinfo{person}{Wengang Zhou}, {and} \bibinfo{person}{Houqiang Li}.} \bibinfo{year}{2021}\natexlab{}.
\newblock \showarticletitle{Transvg: End-to-end visual grounding with transformers}. In \bibinfo{booktitle}{\emph{Proceedings of the IEEE/CVF International Conference on Computer Vision}}. \bibinfo{pages}{1769--1779}.
\newblock


\bibitem[Deruyttere et~al\mbox{.}(2019)]%
        {deruyttere2019talk2car}
\bibfield{author}{\bibinfo{person}{Thierry Deruyttere}, \bibinfo{person}{Simon Vandenhende}, \bibinfo{person}{Dusan Grujicic}, \bibinfo{person}{Luc Van~Gool}, {and} \bibinfo{person}{Marie~Francine Moens}.} \bibinfo{year}{2019}\natexlab{}.
\newblock \showarticletitle{Talk2Car: Taking Control of Your Self-Driving Car}. In \bibinfo{booktitle}{\emph{Proceedings of the 2019 Conference on Empirical Methods in Natural Language Processing and the 9th International Joint Conference on Natural Language Processing (EMNLP-IJCNLP)}}. \bibinfo{pages}{2088--2098}.
\newblock


\bibitem[Ding et~al\mbox{.}(2022)]%
        {ding2022vlt}
\bibfield{author}{\bibinfo{person}{Henghui Ding}, \bibinfo{person}{Chang Liu}, \bibinfo{person}{Suchen Wang}, {and} \bibinfo{person}{Xudong Jiang}.} \bibinfo{year}{2022}\natexlab{}.
\newblock \showarticletitle{Vlt: Vision-language transformer and query generation for referring segmentation}.
\newblock \bibinfo{journal}{\emph{IEEE Transactions on Pattern Analysis and Machine Intelligence}} (\bibinfo{year}{2022}).
\newblock


\bibitem[Dosovitskiy et~al\mbox{.}(2020)]%
        {dosovitskiy2020image}
\bibfield{author}{\bibinfo{person}{Alexey Dosovitskiy}, \bibinfo{person}{Lucas Beyer}, \bibinfo{person}{Alexander Kolesnikov}, \bibinfo{person}{Dirk Weissenborn}, \bibinfo{person}{Xiaohua Zhai}, \bibinfo{person}{Thomas Unterthiner}, \bibinfo{person}{Mostafa Dehghani}, \bibinfo{person}{Matthias Minderer}, \bibinfo{person}{Georg Heigold}, \bibinfo{person}{Sylvain Gelly}, {et~al\mbox{.}}} \bibinfo{year}{2020}\natexlab{}.
\newblock \showarticletitle{An Image is Worth 16x16 Words: Transformers for Image Recognition at Scale}. In \bibinfo{booktitle}{\emph{International Conference on Learning Representations}}.
\newblock


\bibitem[Du et~al\mbox{.}(2022)]%
        {du2022visual}
\bibfield{author}{\bibinfo{person}{Ye Du}, \bibinfo{person}{Zehua Fu}, \bibinfo{person}{Qingjie Liu}, {and} \bibinfo{person}{Yunhong Wang}.} \bibinfo{year}{2022}\natexlab{}.
\newblock \showarticletitle{Visual grounding with transformers}. In \bibinfo{booktitle}{\emph{2022 IEEE International Conference on Multimedia and Expo (ICME)}}. IEEE, \bibinfo{pages}{1--6}.
\newblock


\bibitem[Guan et~al\mbox{.}(2023a)]%
        {runwei2023literec}
\bibfield{author}{\bibinfo{person}{Runwei Guan}, \bibinfo{person}{Ka~Lok Man}, \bibinfo{person}{Eng~Gee Lim}, \bibinfo{person}{Jeremy Smith}, \bibinfo{person}{Yutao Yue}, {and} \bibinfo{person}{Weiping Ding}.} \bibinfo{year}{2023}\natexlab{a}.
\newblock \showarticletitle{Lightweight Traffic Referring Expression Comprehension System Based on Sparse Cross-Modal Attention and Multi-Modal Joint Augmentation}.
\newblock  (\bibinfo{year}{2023}).
\newblock
\urldef\tempurl%
\url{https://doi.org/10.13140/RG.2.2.28608.51204}
\showDOI{\tempurl}


\bibitem[Guan et~al\mbox{.}(2024)]%
        {guan2024mask}
\bibfield{author}{\bibinfo{person}{Runwei Guan}, \bibinfo{person}{Shanliang Yao}, \bibinfo{person}{Lulu Liu}, \bibinfo{person}{Xiaohui Zhu}, \bibinfo{person}{Ka~Lok Man}, \bibinfo{person}{Yong Yue}, \bibinfo{person}{Jeremy Smith}, \bibinfo{person}{Eng~Gee Lim}, {and} \bibinfo{person}{Yutao Yue}.} \bibinfo{year}{2024}\natexlab{}.
\newblock \showarticletitle{Mask-VRDet: A robust riverway panoptic perception model based on dual graph fusion of vision and 4D mmWave radar}.
\newblock \bibinfo{journal}{\emph{Robotics and Autonomous Systems}}  \bibinfo{volume}{171} (\bibinfo{year}{2024}), \bibinfo{pages}{104572}.
\newblock


\bibitem[Guan et~al\mbox{.}(2023b)]%
        {guan2023achelous}
\bibfield{author}{\bibinfo{person}{Runwei Guan}, \bibinfo{person}{Shanliang Yao}, \bibinfo{person}{Xiaohui Zhu}, \bibinfo{person}{Ka~Lok Man}, \bibinfo{person}{Eng~Gee Lim}, \bibinfo{person}{Jeremy Smith}, \bibinfo{person}{Yong Yue}, {and} \bibinfo{person}{Yutao Yue}.} \bibinfo{year}{2023}\natexlab{b}.
\newblock \showarticletitle{Achelous: A fast unified water-surface panoptic perception framework based on fusion of monocular camera and 4d mmwave radar}. In \bibinfo{booktitle}{\emph{2023 IEEE 26th International Conference on Intelligent Transportation Systems (ITSC)}}. IEEE, \bibinfo{pages}{182--188}.
\newblock


\bibitem[Guan et~al\mbox{.}(2023c)]%
        {guan2023efficient}
\bibfield{author}{\bibinfo{person}{Runwei Guan}, \bibinfo{person}{Shanliang Yao}, \bibinfo{person}{Xiaohui Zhu}, \bibinfo{person}{Ka~Lok Man}, \bibinfo{person}{Yong Yue}, \bibinfo{person}{Jeremy Smith}, \bibinfo{person}{Eng~Gee Lim}, {and} \bibinfo{person}{Yutao Yue}.} \bibinfo{year}{2023}\natexlab{c}.
\newblock \showarticletitle{Efficient-vrnet: An exquisite fusion network for riverway panoptic perception based on asymmetric fair fusion of vision and 4d mmwave radar}.
\newblock \bibinfo{journal}{\emph{arXiv preprint arXiv:2308.10287}} (\bibinfo{year}{2023}).
\newblock


\bibitem[Guan et~al\mbox{.}(2023d)]%
        {guan2023achelous++}
\bibfield{author}{\bibinfo{person}{Runwei Guan}, \bibinfo{person}{Haocheng Zhao}, \bibinfo{person}{Shanliang Yao}, \bibinfo{person}{Ka~Lok Man}, \bibinfo{person}{Xiaohui Zhu}, \bibinfo{person}{Limin Yu}, \bibinfo{person}{Yong Yue}, \bibinfo{person}{Jeremy Smith}, \bibinfo{person}{Eng~Gee Lim}, \bibinfo{person}{Weiping Ding}, {et~al\mbox{.}}} \bibinfo{year}{2023}\natexlab{d}.
\newblock \showarticletitle{Achelous++: Power-Oriented Water-Surface Panoptic Perception Framework on Edge Devices based on Vision-Radar Fusion and Pruning of Heterogeneous Modalities}.
\newblock \bibinfo{journal}{\emph{arXiv preprint arXiv:2312.08851}} (\bibinfo{year}{2023}).
\newblock


\bibitem[Guo et~al\mbox{.}(2023)]%
        {guo2023asynchronous}
\bibfield{author}{\bibinfo{person}{Yu Guo}, \bibinfo{person}{Ryan~Wen Liu}, \bibinfo{person}{Jingxiang Qu}, \bibinfo{person}{Yuxu Lu}, \bibinfo{person}{Fenghua Zhu}, {and} \bibinfo{person}{Yisheng Lv}.} \bibinfo{year}{2023}\natexlab{}.
\newblock \showarticletitle{Asynchronous Trajectory Matching-Based Multimodal Maritime Data Fusion for Vessel Traffic Surveillance in Inland Waterways}.
\newblock \bibinfo{journal}{\emph{IEEE Transactions on Intelligent Transportation Systems}} (\bibinfo{year}{2023}).
\newblock


\bibitem[Han et~al\mbox{.}(2023)]%
        {han2023agent}
\bibfield{author}{\bibinfo{person}{Dongchen Han}, \bibinfo{person}{Tianzhu Ye}, \bibinfo{person}{Yizeng Han}, \bibinfo{person}{Zhuofan Xia}, \bibinfo{person}{Shiji Song}, {and} \bibinfo{person}{Gao Huang}.} \bibinfo{year}{2023}\natexlab{}.
\newblock \showarticletitle{Agent Attention: On the Integration of Softmax and Linear Attention}.
\newblock \bibinfo{journal}{\emph{arXiv preprint arXiv:2312.08874}} (\bibinfo{year}{2023}).
\newblock


\bibitem[He et~al\mbox{.}(2023)]%
        {he2023grec}
\bibfield{author}{\bibinfo{person}{Shuting He}, \bibinfo{person}{Henghui Ding}, \bibinfo{person}{Chang Liu}, {and} \bibinfo{person}{Xudong Jiang}.} \bibinfo{year}{2023}\natexlab{}.
\newblock \showarticletitle{Grec: Generalized referring expression comprehension}.
\newblock \bibinfo{journal}{\emph{arXiv preprint arXiv:2308.16182}} (\bibinfo{year}{2023}).
\newblock


\bibitem[Hong et~al\mbox{.}(2019)]%
        {hong2019learning}
\bibfield{author}{\bibinfo{person}{Richang Hong}, \bibinfo{person}{Daqing Liu}, \bibinfo{person}{Xiaoyu Mo}, \bibinfo{person}{Xiangnan He}, {and} \bibinfo{person}{Hanwang Zhang}.} \bibinfo{year}{2019}\natexlab{}.
\newblock \showarticletitle{Learning to compose and reason with language tree structures for visual grounding}.
\newblock \bibinfo{journal}{\emph{IEEE transactions on pattern analysis and machine intelligence}} \bibinfo{volume}{44}, \bibinfo{number}{2} (\bibinfo{year}{2019}), \bibinfo{pages}{684--696}.
\newblock


\bibitem[Jocher et~al\mbox{.}(2023)]%
        {Jocher_Ultralytics_YOLO_2023}
\bibfield{author}{\bibinfo{person}{Glenn Jocher}, \bibinfo{person}{Ayush Chaurasia}, {and} \bibinfo{person}{Jing Qiu}.} \bibinfo{year}{2023}\natexlab{}.
\newblock \bibinfo{booktitle}{\emph{{Ultralytics YOLO}}}.
\newblock
\urldef\tempurl%
\url{https://github.com/ultralytics/ultralytics}
\showURL{%
\tempurl}


\bibitem[Kazemzadeh et~al\mbox{.}(2014)]%
        {kazemzadeh2014referitgame}
\bibfield{author}{\bibinfo{person}{Sahar Kazemzadeh}, \bibinfo{person}{Vicente Ordonez}, \bibinfo{person}{Mark Matten}, {and} \bibinfo{person}{Tamara Berg}.} \bibinfo{year}{2014}\natexlab{}.
\newblock \showarticletitle{Referitgame: Referring to objects in photographs of natural scenes}. In \bibinfo{booktitle}{\emph{Proceedings of the 2014 conference on empirical methods in natural language processing (EMNLP)}}. \bibinfo{pages}{787--798}.
\newblock


\bibitem[Kendall et~al\mbox{.}(2018)]%
        {kendall2018multi}
\bibfield{author}{\bibinfo{person}{Alex Kendall}, \bibinfo{person}{Yarin Gal}, {and} \bibinfo{person}{Roberto Cipolla}.} \bibinfo{year}{2018}\natexlab{}.
\newblock \showarticletitle{Multi-task learning using uncertainty to weigh losses for scene geometry and semantics}. In \bibinfo{booktitle}{\emph{Proceedings of the IEEE conference on computer vision and pattern recognition}}. \bibinfo{pages}{7482--7491}.
\newblock


\bibitem[Krishna et~al\mbox{.}(2017)]%
        {krishna2017visual}
\bibfield{author}{\bibinfo{person}{Ranjay Krishna}, \bibinfo{person}{Yuke Zhu}, \bibinfo{person}{Oliver Groth}, \bibinfo{person}{Justin Johnson}, \bibinfo{person}{Kenji Hata}, \bibinfo{person}{Joshua Kravitz}, \bibinfo{person}{Stephanie Chen}, \bibinfo{person}{Yannis Kalantidis}, \bibinfo{person}{Li-Jia Li}, \bibinfo{person}{David~A Shamma}, {et~al\mbox{.}}} \bibinfo{year}{2017}\natexlab{}.
\newblock \showarticletitle{Visual genome: Connecting language and vision using crowdsourced dense image annotations}.
\newblock \bibinfo{journal}{\emph{International journal of computer vision}}  \bibinfo{volume}{123} (\bibinfo{year}{2017}), \bibinfo{pages}{32--73}.
\newblock


\bibitem[Kristan et~al\mbox{.}(2015)]%
        {kristan2015fast}
\bibfield{author}{\bibinfo{person}{Matej Kristan}, \bibinfo{person}{Vildana~Suli{\'c} Kenk}, \bibinfo{person}{Stanislav Kova{\v{c}}i{\v{c}}}, {and} \bibinfo{person}{Janez Per{\v{s}}}.} \bibinfo{year}{2015}\natexlab{}.
\newblock \showarticletitle{Fast image-based obstacle detection from unmanned surface vehicles}.
\newblock \bibinfo{journal}{\emph{IEEE transactions on cybernetics}} \bibinfo{volume}{46}, \bibinfo{number}{3} (\bibinfo{year}{2015}), \bibinfo{pages}{641--654}.
\newblock


\bibitem[Lan et~al\mbox{.}(2020)]%
        {lan2020albert}
\bibfield{author}{\bibinfo{person}{Zhenzhong Lan}, \bibinfo{person}{Mingda Chen}, \bibinfo{person}{Sebastian Goodman}, \bibinfo{person}{Kevin Gimpel}, \bibinfo{person}{Piyush Sharma}, {and} \bibinfo{person}{Radu Soricut}.} \bibinfo{year}{2020}\natexlab{}.
\newblock \showarticletitle{ALBERT: A Lite BERT for Self-supervised Learning of Language Representations}.
\newblock  (\bibinfo{year}{2020}).
\newblock


\bibitem[Li et~al\mbox{.}(2022)]%
        {li2022yolov6}
\bibfield{author}{\bibinfo{person}{Chuyi Li}, \bibinfo{person}{Lulu Li}, \bibinfo{person}{Hongliang Jiang}, \bibinfo{person}{Kaiheng Weng}, \bibinfo{person}{Yifei Geng}, \bibinfo{person}{Liang Li}, \bibinfo{person}{Zaidan Ke}, \bibinfo{person}{Qingyuan Li}, \bibinfo{person}{Meng Cheng}, \bibinfo{person}{Weiqiang Nie}, {et~al\mbox{.}}} \bibinfo{year}{2022}\natexlab{}.
\newblock \showarticletitle{YOLOv6: A single-stage object detection framework for industrial applications}.
\newblock \bibinfo{journal}{\emph{arXiv preprint arXiv:2209.02976}} (\bibinfo{year}{2022}).
\newblock


\bibitem[Li and Sigal(2021)]%
        {li2021referring}
\bibfield{author}{\bibinfo{person}{Muchen Li} {and} \bibinfo{person}{Leonid Sigal}.} \bibinfo{year}{2021}\natexlab{}.
\newblock \showarticletitle{Referring transformer: A one-step approach to multi-task visual grounding}.
\newblock \bibinfo{journal}{\emph{Advances in neural information processing systems}}  \bibinfo{volume}{34} (\bibinfo{year}{2021}), \bibinfo{pages}{19652--19664}.
\newblock


\bibitem[Li et~al\mbox{.}(2020)]%
        {li2020generalized}
\bibfield{author}{\bibinfo{person}{Xiang Li}, \bibinfo{person}{Wenhai Wang}, \bibinfo{person}{Lijun Wu}, \bibinfo{person}{Shuo Chen}, \bibinfo{person}{Xiaolin Hu}, \bibinfo{person}{Jun Li}, \bibinfo{person}{Jinhui Tang}, {and} \bibinfo{person}{Jian Yang}.} \bibinfo{year}{2020}\natexlab{}.
\newblock \showarticletitle{Generalized focal loss: Learning qualified and distributed bounding boxes for dense object detection}.
\newblock \bibinfo{journal}{\emph{Advances in Neural Information Processing Systems}}  \bibinfo{volume}{33} (\bibinfo{year}{2020}), \bibinfo{pages}{21002--21012}.
\newblock


\bibitem[Lin et~al\mbox{.}(2017)]%
        {lin2017focal}
\bibfield{author}{\bibinfo{person}{Tsung-Yi Lin}, \bibinfo{person}{Priya Goyal}, \bibinfo{person}{Ross Girshick}, \bibinfo{person}{Kaiming He}, {and} \bibinfo{person}{Piotr Doll{\'a}r}.} \bibinfo{year}{2017}\natexlab{}.
\newblock \showarticletitle{Focal loss for dense object detection}. In \bibinfo{booktitle}{\emph{Proceedings of the IEEE international conference on computer vision}}. \bibinfo{pages}{2980--2988}.
\newblock


\bibitem[Liu et~al\mbox{.}(2023)]%
        {liu2023gres}
\bibfield{author}{\bibinfo{person}{Chang Liu}, \bibinfo{person}{Henghui Ding}, {and} \bibinfo{person}{Xudong Jiang}.} \bibinfo{year}{2023}\natexlab{}.
\newblock \showarticletitle{GRES: Generalized referring expression segmentation}. In \bibinfo{booktitle}{\emph{Proceedings of the IEEE/CVF Conference on Computer Vision and Pattern Recognition}}. \bibinfo{pages}{23592--23601}.
\newblock


\bibitem[Liu et~al\mbox{.}(2019)]%
        {liu2019learning}
\bibfield{author}{\bibinfo{person}{Daqing Liu}, \bibinfo{person}{Hanwang Zhang}, \bibinfo{person}{Feng Wu}, {and} \bibinfo{person}{Zheng-Jun Zha}.} \bibinfo{year}{2019}\natexlab{}.
\newblock \showarticletitle{Learning to assemble neural module tree networks for visual grounding}. In \bibinfo{booktitle}{\emph{Proceedings of the IEEE/CVF International Conference on Computer Vision}}. \bibinfo{pages}{4673--4682}.
\newblock


\bibitem[Liu et~al\mbox{.}(2021)]%
        {liu2021efficient}
\bibfield{author}{\bibinfo{person}{Jingyi Liu}, \bibinfo{person}{Hengyu Li}, \bibinfo{person}{Jun Luo}, \bibinfo{person}{Shaorong Xie}, {and} \bibinfo{person}{Yu Sun}.} \bibinfo{year}{2021}\natexlab{}.
\newblock \showarticletitle{Efficient obstacle detection based on prior estimation network and spatially constrained mixture model for unmanned surface vehicles}.
\newblock \bibinfo{journal}{\emph{Journal of Field Robotics}} \bibinfo{volume}{38}, \bibinfo{number}{2} (\bibinfo{year}{2021}), \bibinfo{pages}{212--228}.
\newblock


\bibitem[Luo et~al\mbox{.}(2022)]%
        {luo2022what}
\bibfield{author}{\bibinfo{person}{Gen Luo}, \bibinfo{person}{Yiyi Zhou}, \bibinfo{person}{Jiamu Sun}, \bibinfo{person}{Shubin Huang}, \bibinfo{person}{Xiaoshuai Sun}, \bibinfo{person}{Qixiang Ye}, \bibinfo{person}{Yongjian Wu}, {and} \bibinfo{person}{Rongrong Ji}.} \bibinfo{year}{2022}\natexlab{}.
\newblock \showarticletitle{What Goes beyond Multi-modal Fusion in One-stage Referring Expression Comprehension: An Empirical Study}.
\newblock \bibinfo{journal}{\emph{arXiv preprint arXiv:2204.07913}} (\bibinfo{year}{2022}).
\newblock


\bibitem[Luo et~al\mbox{.}(2020)]%
        {luo2020multi}
\bibfield{author}{\bibinfo{person}{Gen Luo}, \bibinfo{person}{Yiyi Zhou}, \bibinfo{person}{Xiaoshuai Sun}, \bibinfo{person}{Liujuan Cao}, \bibinfo{person}{Chenglin Wu}, \bibinfo{person}{Cheng Deng}, {and} \bibinfo{person}{Rongrong Ji}.} \bibinfo{year}{2020}\natexlab{}.
\newblock \showarticletitle{Multi-task collaborative network for joint referring expression comprehension and segmentation}. In \bibinfo{booktitle}{\emph{Proceedings of the IEEE/CVF Conference on computer vision and pattern recognition}}. \bibinfo{pages}{10034--10043}.
\newblock


\bibitem[Miller et~al\mbox{.}(2018)]%
        {miller2018visual}
\bibfield{author}{\bibinfo{person}{Martin Miller}, \bibinfo{person}{Soon-Jo Chung}, {and} \bibinfo{person}{Seth Hutchinson}.} \bibinfo{year}{2018}\natexlab{}.
\newblock \showarticletitle{The visual--inertial canoe dataset}.
\newblock \bibinfo{journal}{\emph{The International Journal of Robotics Research}} \bibinfo{volume}{37}, \bibinfo{number}{1} (\bibinfo{year}{2018}), \bibinfo{pages}{13--20}.
\newblock


\bibitem[Moosbauer et~al\mbox{.}(2019)]%
        {moosbauer2019benchmark}
\bibfield{author}{\bibinfo{person}{Sebastian Moosbauer}, \bibinfo{person}{Daniel Konig}, \bibinfo{person}{Jens Jakel}, {and} \bibinfo{person}{Michael Teutsch}.} \bibinfo{year}{2019}\natexlab{}.
\newblock \showarticletitle{A benchmark for deep learning based object detection in maritime environments}. In \bibinfo{booktitle}{\emph{Proceedings of the IEEE/CVF conference on computer vision and pattern recognition workshops}}. \bibinfo{pages}{0--0}.
\newblock


\bibitem[Ouyang et~al\mbox{.}(2022)]%
        {ouyang2022training}
\bibfield{author}{\bibinfo{person}{Long Ouyang}, \bibinfo{person}{Jeffrey Wu}, \bibinfo{person}{Xu Jiang}, \bibinfo{person}{Diogo Almeida}, \bibinfo{person}{Carroll Wainwright}, \bibinfo{person}{Pamela Mishkin}, \bibinfo{person}{Chong Zhang}, \bibinfo{person}{Sandhini Agarwal}, \bibinfo{person}{Katarina Slama}, \bibinfo{person}{Alex Ray}, {et~al\mbox{.}}} \bibinfo{year}{2022}\natexlab{}.
\newblock \showarticletitle{Training language models to follow instructions with human feedback}.
\newblock \bibinfo{journal}{\emph{Advances in neural information processing systems}}  \bibinfo{volume}{35} (\bibinfo{year}{2022}), \bibinfo{pages}{27730--27744}.
\newblock


\bibitem[Peng et~al\mbox{.}(2022)]%
        {peng2022pp}
\bibfield{author}{\bibinfo{person}{Juncai Peng}, \bibinfo{person}{Yi Liu}, \bibinfo{person}{Shiyu Tang}, \bibinfo{person}{Yuying Hao}, \bibinfo{person}{Lutao Chu}, \bibinfo{person}{Guowei Chen}, \bibinfo{person}{Zewu Wu}, \bibinfo{person}{Zeyu Chen}, \bibinfo{person}{Zhiliang Yu}, \bibinfo{person}{Yuning Du}, {et~al\mbox{.}}} \bibinfo{year}{2022}\natexlab{}.
\newblock \showarticletitle{Pp-liteseg: A superior real-time semantic segmentation model}.
\newblock \bibinfo{journal}{\emph{arXiv preprint arXiv:2204.02681}} (\bibinfo{year}{2022}).
\newblock


\bibitem[Plummer et~al\mbox{.}(2015)]%
        {plummer2015flickr30k}
\bibfield{author}{\bibinfo{person}{Bryan~A Plummer}, \bibinfo{person}{Liwei Wang}, \bibinfo{person}{Chris~M Cervantes}, \bibinfo{person}{Juan~C Caicedo}, \bibinfo{person}{Julia Hockenmaier}, {and} \bibinfo{person}{Svetlana Lazebnik}.} \bibinfo{year}{2015}\natexlab{}.
\newblock \showarticletitle{Flickr30k entities: Collecting region-to-phrase correspondences for richer image-to-sentence models}. In \bibinfo{booktitle}{\emph{Proceedings of the IEEE international conference on computer vision}}. \bibinfo{pages}{2641--2649}.
\newblock


\bibitem[Qiao et~al\mbox{.}(2020)]%
        {qiao2020referring}
\bibfield{author}{\bibinfo{person}{Yanyuan Qiao}, \bibinfo{person}{Chaorui Deng}, {and} \bibinfo{person}{Qi Wu}.} \bibinfo{year}{2020}\natexlab{}.
\newblock \showarticletitle{Referring expression comprehension: A survey of methods and datasets}.
\newblock \bibinfo{journal}{\emph{IEEE Transactions on Multimedia}}  \bibinfo{volume}{23} (\bibinfo{year}{2020}), \bibinfo{pages}{4426--4440}.
\newblock


\bibitem[Sudre et~al\mbox{.}(2017)]%
        {sudre2017generalised}
\bibfield{author}{\bibinfo{person}{Carole~H Sudre}, \bibinfo{person}{Wenqi Li}, \bibinfo{person}{Tom Vercauteren}, \bibinfo{person}{Sebastien Ourselin}, {and} \bibinfo{person}{M Jorge~Cardoso}.} \bibinfo{year}{2017}\natexlab{}.
\newblock \showarticletitle{Generalised dice overlap as a deep learning loss function for highly unbalanced segmentations}. In \bibinfo{booktitle}{\emph{Deep Learning in Medical Image Analysis and Multimodal Learning for Clinical Decision Support: Third International Workshop, DLMIA 2017, and 7th International Workshop, ML-CDS 2017, Held in Conjunction with MICCAI 2017, Qu{\'e}bec City, QC, Canada, September 14, Proceedings 3}}. Springer, \bibinfo{pages}{240--248}.
\newblock


\bibitem[Taipalmaa et~al\mbox{.}(2019)]%
        {taipalmaa2019high}
\bibfield{author}{\bibinfo{person}{Jussi Taipalmaa}, \bibinfo{person}{Nikolaos Passalis}, \bibinfo{person}{Honglei Zhang}, \bibinfo{person}{Moncef Gabbouj}, {and} \bibinfo{person}{Jenni Raitoharju}.} \bibinfo{year}{2019}\natexlab{}.
\newblock \showarticletitle{High-resolution water segmentation for autonomous unmanned surface vehicles: A novel dataset and evaluation}. In \bibinfo{booktitle}{\emph{2019 IEEE 29th International Workshop on Machine Learning for Signal Processing (MLSP)}}. IEEE, \bibinfo{pages}{1--6}.
\newblock


\bibitem[Tu et~al\mbox{.}(2023)]%
        {tu2023femtodet}
\bibfield{author}{\bibinfo{person}{Peng Tu}, \bibinfo{person}{Xu Xie}, \bibinfo{person}{Guo Ai}, \bibinfo{person}{Yuexiang Li}, \bibinfo{person}{Yawen Huang}, {and} \bibinfo{person}{Yefeng Zheng}.} \bibinfo{year}{2023}\natexlab{}.
\newblock \showarticletitle{Femtodet: An object detection baseline for energy versus performance tradeoffs}. In \bibinfo{booktitle}{\emph{Proceedings of the IEEE/CVF International Conference on Computer Vision}}. \bibinfo{pages}{13318--13327}.
\newblock


\bibitem[Varga et~al\mbox{.}(2022)]%
        {varga2022seadronessee}
\bibfield{author}{\bibinfo{person}{Leon~Amadeus Varga}, \bibinfo{person}{Benjamin Kiefer}, \bibinfo{person}{Martin Messmer}, {and} \bibinfo{person}{Andreas Zell}.} \bibinfo{year}{2022}\natexlab{}.
\newblock \showarticletitle{Seadronessee: A maritime benchmark for detecting humans in open water}. In \bibinfo{booktitle}{\emph{Proceedings of the IEEE/CVF winter conference on applications of computer vision}}. \bibinfo{pages}{2260--2270}.
\newblock


\bibitem[Vasu et~al\mbox{.}(2023)]%
        {vasu2023fastvit}
\bibfield{author}{\bibinfo{person}{Pavan Kumar~Anasosalu Vasu}, \bibinfo{person}{James Gabriel}, \bibinfo{person}{Jeff Zhu}, \bibinfo{person}{Oncel Tuzel}, {and} \bibinfo{person}{Anurag Ranjan}.} \bibinfo{year}{2023}\natexlab{}.
\newblock \showarticletitle{FastViT: A fast hybrid vision transformer using structural reparameterization}. In \bibinfo{booktitle}{\emph{Proceedings of the IEEE/CVF International Conference on Computer Vision}}. \bibinfo{pages}{5785--5795}.
\newblock


\bibitem[Vaswani et~al\mbox{.}(2017)]%
        {vaswani2017attention}
\bibfield{author}{\bibinfo{person}{Ashish Vaswani}, \bibinfo{person}{Noam Shazeer}, \bibinfo{person}{Niki Parmar}, \bibinfo{person}{Jakob Uszkoreit}, \bibinfo{person}{Llion Jones}, \bibinfo{person}{Aidan~N Gomez}, \bibinfo{person}{{\L}ukasz Kaiser}, {and} \bibinfo{person}{Illia Polosukhin}.} \bibinfo{year}{2017}\natexlab{}.
\newblock \showarticletitle{Attention is all you need}.
\newblock \bibinfo{journal}{\emph{Advances in neural information processing systems}}  \bibinfo{volume}{30} (\bibinfo{year}{2017}).
\newblock


\bibitem[Wang et~al\mbox{.}(2020)]%
        {wang2020give}
\bibfield{author}{\bibinfo{person}{Peng Wang}, \bibinfo{person}{Dongyang Liu}, \bibinfo{person}{Hui Li}, {and} \bibinfo{person}{Qi Wu}.} \bibinfo{year}{2020}\natexlab{}.
\newblock \showarticletitle{Give me something to eat: Referring expression comprehension with commonsense knowledge}. In \bibinfo{booktitle}{\emph{Proceedings of the 28th ACM International Conference on Multimedia}}. \bibinfo{pages}{28--36}.
\newblock


\bibitem[Wang et~al\mbox{.}(2019)]%
        {wang2019neighbourhood}
\bibfield{author}{\bibinfo{person}{Peng Wang}, \bibinfo{person}{Qi Wu}, \bibinfo{person}{Jiewei Cao}, \bibinfo{person}{Chunhua Shen}, \bibinfo{person}{Lianli Gao}, {and} \bibinfo{person}{Anton van~den Hengel}.} \bibinfo{year}{2019}\natexlab{}.
\newblock \showarticletitle{Neighbourhood watch: Referring expression comprehension via language-guided graph attention networks}. In \bibinfo{booktitle}{\emph{Proceedings of the IEEE/CVF Conference on Computer Vision and Pattern Recognition}}. \bibinfo{pages}{1960--1968}.
\newblock


\bibitem[Wu et~al\mbox{.}(2023)]%
        {wu2023referring}
\bibfield{author}{\bibinfo{person}{Dongming Wu}, \bibinfo{person}{Wencheng Han}, \bibinfo{person}{Tiancai Wang}, \bibinfo{person}{Xingping Dong}, \bibinfo{person}{Xiangyu Zhang}, {and} \bibinfo{person}{Jianbing Shen}.} \bibinfo{year}{2023}\natexlab{}.
\newblock \showarticletitle{Referring Multi-Object Tracking}. In \bibinfo{booktitle}{\emph{Proceedings of the IEEE/CVF Conference on Computer Vision and Pattern Recognition}}. \bibinfo{pages}{14633--14642}.
\newblock


\bibitem[Yang et~al\mbox{.}(2019b)]%
        {yang2019dynamic}
\bibfield{author}{\bibinfo{person}{Sibei Yang}, \bibinfo{person}{Guanbin Li}, {and} \bibinfo{person}{Yizhou Yu}.} \bibinfo{year}{2019}\natexlab{b}.
\newblock \showarticletitle{Dynamic graph attention for referring expression comprehension}. In \bibinfo{booktitle}{\emph{Proceedings of the IEEE/CVF International Conference on Computer Vision}}. \bibinfo{pages}{4644--4653}.
\newblock


\bibitem[Yang et~al\mbox{.}(2019a)]%
        {yang2019fast}
\bibfield{author}{\bibinfo{person}{Zhengyuan Yang}, \bibinfo{person}{Boqing Gong}, \bibinfo{person}{Liwei Wang}, \bibinfo{person}{Wenbing Huang}, \bibinfo{person}{Dong Yu}, {and} \bibinfo{person}{Jiebo Luo}.} \bibinfo{year}{2019}\natexlab{a}.
\newblock \showarticletitle{A fast and accurate one-stage approach to visual grounding}. In \bibinfo{booktitle}{\emph{Proceedings of the IEEE/CVF International Conference on Computer Vision}}. \bibinfo{pages}{4683--4693}.
\newblock


\bibitem[Yao et~al\mbox{.}(2023a)]%
        {yaoreview2023}
\bibfield{author}{\bibinfo{person}{Shanliang Yao}, \bibinfo{person}{Runwei Guan}, \bibinfo{person}{Xiaoyu Huang}, \bibinfo{person}{Zhuoxiao Li}, \bibinfo{person}{Xiangyu Sha}, \bibinfo{person}{Yong Yue}, \bibinfo{person}{Eng~Gee Lim}, \bibinfo{person}{Hyungjoon Seo}, \bibinfo{person}{Ka~Lok Man}, \bibinfo{person}{Xiaohui Zhu}, {and} \bibinfo{person}{Yutao Yue}.} \bibinfo{year}{2023}\natexlab{a}.
\newblock \showarticletitle{Radar-Camera Fusion for Object Detection and Semantic Segmentation in Autonomous Driving: A Comprehensive Review}.
\newblock \bibinfo{journal}{\emph{IEEE Transactions on Intelligent Vehicles}} (\bibinfo{year}{2023}), \bibinfo{pages}{1--40}.
\newblock
\urldef\tempurl%
\url{https://doi.org/10.1109/TIV.2023.3307157}
\showDOI{\tempurl}


\bibitem[Yao et~al\mbox{.}(2023b)]%
        {yao2023radar}
\bibfield{author}{\bibinfo{person}{Shanliang Yao}, \bibinfo{person}{Runwei Guan}, \bibinfo{person}{Zitian Peng}, \bibinfo{person}{Chenhang Xu}, \bibinfo{person}{Yilu Shi}, \bibinfo{person}{Yong Yue}, \bibinfo{person}{Eng~Gee Lim}, \bibinfo{person}{Hyungjoon Seo}, \bibinfo{person}{Ka~Lok Man}, \bibinfo{person}{Xiaohui Zhu}, {et~al\mbox{.}}} \bibinfo{year}{2023}\natexlab{b}.
\newblock \showarticletitle{Radar Perception in Autonomous Driving: Exploring Different Data Representations}.
\newblock \bibinfo{journal}{\emph{arXiv preprint arXiv:2312.04861}} (\bibinfo{year}{2023}).
\newblock


\bibitem[Yao et~al\mbox{.}(2023c)]%
        {yao2023waterscenes}
\bibfield{author}{\bibinfo{person}{Shanliang Yao}, \bibinfo{person}{Runwei Guan}, \bibinfo{person}{Zhaodong Wu}, \bibinfo{person}{Yi Ni}, \bibinfo{person}{Zixian Zhang}, \bibinfo{person}{Zile Huang}, \bibinfo{person}{Xiaohui Zhu}, \bibinfo{person}{Yutao Yue}, \bibinfo{person}{Yong Yue}, \bibinfo{person}{Hyungjoon Seo}, {et~al\mbox{.}}} \bibinfo{year}{2023}\natexlab{c}.
\newblock \showarticletitle{Waterscenes: A multi-task 4d radar-camera fusion dataset and benchmark for autonomous driving on water surfaces}.
\newblock \bibinfo{journal}{\emph{arXiv preprint arXiv:2307.06505}} (\bibinfo{year}{2023}).
\newblock


\bibitem[Yu et~al\mbox{.}(2018)]%
        {yu2018mattnet}
\bibfield{author}{\bibinfo{person}{Licheng Yu}, \bibinfo{person}{Zhe Lin}, \bibinfo{person}{Xiaohui Shen}, \bibinfo{person}{Jimei Yang}, \bibinfo{person}{Xin Lu}, \bibinfo{person}{Mohit Bansal}, {and} \bibinfo{person}{Tamara~L Berg}.} \bibinfo{year}{2018}\natexlab{}.
\newblock \showarticletitle{Mattnet: Modular attention network for referring expression comprehension}. In \bibinfo{booktitle}{\emph{Proceedings of the IEEE conference on computer vision and pattern recognition}}. \bibinfo{pages}{1307--1315}.
\newblock


\bibitem[Yu et~al\mbox{.}(2016)]%
        {yu2016modeling}
\bibfield{author}{\bibinfo{person}{Licheng Yu}, \bibinfo{person}{Patrick Poirson}, \bibinfo{person}{Shan Yang}, \bibinfo{person}{Alexander~C Berg}, {and} \bibinfo{person}{Tamara~L Berg}.} \bibinfo{year}{2016}\natexlab{}.
\newblock \showarticletitle{Modeling context in referring expressions}. In \bibinfo{booktitle}{\emph{Computer Vision--ECCV 2016: 14th European Conference, Amsterdam, The Netherlands, October 11-14, 2016, Proceedings, Part II 14}}. Springer, \bibinfo{pages}{69--85}.
\newblock


\bibitem[Zhan et~al\mbox{.}(2023)]%
        {zhan2023rsvg}
\bibfield{author}{\bibinfo{person}{Yang Zhan}, \bibinfo{person}{Zhitong Xiong}, {and} \bibinfo{person}{Yuan Yuan}.} \bibinfo{year}{2023}\natexlab{}.
\newblock \showarticletitle{Rsvg: Exploring data and models for visual grounding on remote sensing data}.
\newblock \bibinfo{journal}{\emph{IEEE Transactions on Geoscience and Remote Sensing}}  \bibinfo{volume}{61} (\bibinfo{year}{2023}), \bibinfo{pages}{1--13}.
\newblock


\bibitem[Zhang et~al\mbox{.}(2021)]%
        {zhang2021robust}
\bibfield{author}{\bibinfo{person}{Wen Zhang}, \bibinfo{person}{Xujie He}, \bibinfo{person}{Wanyi Li}, \bibinfo{person}{Zhi Zhang}, \bibinfo{person}{Yongkang Luo}, \bibinfo{person}{Li Su}, {and} \bibinfo{person}{Peng Wang}.} \bibinfo{year}{2021}\natexlab{}.
\newblock \showarticletitle{A robust deep affinity network for multiple ship tracking}.
\newblock \bibinfo{journal}{\emph{IEEE Transactions on Instrumentation and Measurement}}  \bibinfo{volume}{70} (\bibinfo{year}{2021}), \bibinfo{pages}{1--20}.
\newblock


\bibitem[Zheng et~al\mbox{.}(2020)]%
        {zheng2020distance}
\bibfield{author}{\bibinfo{person}{Zhaohui Zheng}, \bibinfo{person}{Ping Wang}, \bibinfo{person}{Wei Liu}, \bibinfo{person}{Jinze Li}, \bibinfo{person}{Rongguang Ye}, {and} \bibinfo{person}{Dongwei Ren}.} \bibinfo{year}{2020}\natexlab{}.
\newblock \showarticletitle{Distance-IoU loss: Faster and better learning for bounding box regression}. In \bibinfo{booktitle}{\emph{Proceedings of the AAAI conference on artificial intelligence}}, Vol.~\bibinfo{volume}{34}. \bibinfo{pages}{12993--13000}.
\newblock


\bibitem[Zhou et~al\mbox{.}(2023)]%
        {zhou2023vision}
\bibfield{author}{\bibinfo{person}{Xingcheng Zhou}, \bibinfo{person}{Mingyu Liu}, \bibinfo{person}{Bare~Luka Zagar}, \bibinfo{person}{Ekim Yurtsever}, {and} \bibinfo{person}{Alois~C Knoll}.} \bibinfo{year}{2023}\natexlab{}.
\newblock \showarticletitle{Vision language models in autonomous driving and intelligent transportation systems}.
\newblock \bibinfo{journal}{\emph{arXiv preprint arXiv:2310.14414}} (\bibinfo{year}{2023}).
\newblock


\bibitem[Zhou et~al\mbox{.}(2021)]%
        {zhou2021image}
\bibfield{author}{\bibinfo{person}{Zhiguo Zhou}, \bibinfo{person}{Jiaen Sun}, \bibinfo{person}{Jiabao Yu}, \bibinfo{person}{Kaiyuan Liu}, \bibinfo{person}{Junwei Duan}, \bibinfo{person}{Long Chen}, {and} \bibinfo{person}{CL~Philip Chen}.} \bibinfo{year}{2021}\natexlab{}.
\newblock \showarticletitle{An image-based benchmark dataset and a novel object detector for water surface object detection}.
\newblock \bibinfo{journal}{\emph{Frontiers in Neurorobotics}}  \bibinfo{volume}{15} (\bibinfo{year}{2021}), \bibinfo{pages}{723336}.
\newblock


\bibitem[Zhu et~al\mbox{.}(2022)]%
        {zhu2022seqtr}
\bibfield{author}{\bibinfo{person}{Chaoyang Zhu}, \bibinfo{person}{Yiyi Zhou}, \bibinfo{person}{Yunhang Shen}, \bibinfo{person}{Gen Luo}, \bibinfo{person}{Xingjia Pan}, \bibinfo{person}{Mingbao Lin}, \bibinfo{person}{Chao Chen}, \bibinfo{person}{Liujuan Cao}, \bibinfo{person}{Xiaoshuai Sun}, {and} \bibinfo{person}{Rongrong Ji}.} \bibinfo{year}{2022}\natexlab{}.
\newblock \showarticletitle{Seqtr: A simple yet universal network for visual grounding}. In \bibinfo{booktitle}{\emph{European Conference on Computer Vision}}. Springer, \bibinfo{pages}{598--615}.
\newblock


\bibitem[{\v{Z}}ust et~al\mbox{.}(2023)]%
        {vzust2023lars}
\bibfield{author}{\bibinfo{person}{Lojze {\v{Z}}ust}, \bibinfo{person}{Janez Per{\v{s}}}, {and} \bibinfo{person}{Matej Kristan}.} \bibinfo{year}{2023}\natexlab{}.
\newblock \showarticletitle{Lars: A diverse panoptic maritime obstacle detection dataset and benchmark}. In \bibinfo{booktitle}{\emph{Proceedings of the IEEE/CVF International Conference on Computer Vision}}. \bibinfo{pages}{20304--20314}.
\newblock


\end{thebibliography}

\appendix

\end{document}